\documentclass[10pt,twocolumn,letterpaper]{article}

\usepackage[pagenumbers]{cvpr} % To force page numbers, e.g. for an arXiv version

\usepackage{amsmath}
\usepackage{amssymb}
\usepackage{tocloft}
\usepackage{mathtools}
\usepackage{amsthm}
\usepackage{pifont}
\usepackage{newfloat}
\usepackage{listings}
\usepackage{algorithm}
\usepackage{algorithmic}
\usepackage[accsupp]{axessibility} 
\usepackage{adjustbox}
\usepackage{multirow}
\usepackage{booktabs}
\usepackage{tabularx}
\usepackage{color}
\usepackage{bm}
\usepackage{amsmath}
\usepackage{tikz}
\usepackage{xcolor}
\usepackage{tcolorbox}
\usepackage{wrapfig}
\definecolor{cvprblue}{rgb}{0.21,0.49,0.74}
\usepackage[pagebackref,breaklinks,colorlinks,allcolors=cvprblue]{hyperref}

\def\paperID{9404} % *** Enter the Paper ID here
\def\confName{CVPR}
\def\confYear{2025}

\title{Generative Multimodal Pretraining with Discrete Diffusion Timestep Tokens}

\author{Kaihang Pan$^{1}$\footnotemark[1] \quad Wang Lin$^{1}$\footnotemark[1]\quad  Zhongqi Yue$^{2}$\footnotemark[1] \quad Tenglong Ao$^3$ \quad Liyu Jia$^{2}$ \quad Wei Zhao$^4$ \\
Juncheng Li$^1$\footnotemark[2]\quad\quad Siliang Tang$^1$\quad\quad Hanwang Zhang$^{2}$\\ \small $^1$Zhejiang University, $^2$Nanyang Technological University, $^3$Peking University, $^4$Huawei Singapore Research Center \\
{\tt\small \{kaihangpan, linwanglw, junchengli, siliang\}@zju.edu.cn}, \tt\small aubrey.tenglong.ao@gmail.com
\\ \tt\small \{zhongqi.yue, hanwangzhang\}@ntu.edu.sg, liyu002@e.ntu.edu.sg, zhaowei82@huawei.com}

\begin{document}
\maketitle
\renewcommand{\thefootnote}{\fnsymbol{footnote}} %将脚注符号设置为fnsymbol类型，即特殊符号表示
\footnotetext[1]{Equal Contribution.} %对应脚注[1]
\footnotetext[2]{Juncheng Li is the corresponding author.}
\begin{abstract}

Recent endeavors in Multimodal Large Language Models (MLLMs) aim to unify visual comprehension and generation by combining LLM and diffusion models, the state-of-the-art in each task, respectively. Existing approaches rely on spatial visual tokens, where image patches are encoded and arranged according to a spatial order (e.g., raster scan). However, we show that spatial tokens lack the recursive structure inherent to languages, hence form an impossible language for LLM to master. In this paper, we build a proper visual language by leveraging diffusion timesteps to learn discrete, recursive visual tokens. Our proposed tokens recursively compensate for the progressive attribute loss in noisy images as timesteps increase, enabling the diffusion model to reconstruct the original image at any timestep. This approach allows us to effectively integrate the strengths of LLMs in autoregressive reasoning and diffusion models in precise image generation, achieving seamless multimodal comprehension and generation within a unified framework.  Extensive experiments show that we achieve superior performance for multimodal comprehension and generation simultaneously compared with other MLLMs. 
Project Page: \url{https://DDT-LLaMA.github.io/}.

\end{abstract}
    
\addtocontents{toc}{\protect\setcounter{tocdepth}{-1}}
\section{Introduction}
\label{sec:intro}

Multimodal Large Language Models (MLLMs)~\cite{ge2024seed,wang2024emu3} strive to unify the comprehension and generation of data across various modalities within the same next-token prediction paradigm of LLM~\cite{openai2023chatgpt,touvron2023llama}.
Specifically, given a user query about comprehension---``What kind of dog is in this picture [IMG]'', or generation---``Turning the picture [IMG] into a sketch'', the model can complete the task by sequentially predicting the appropriate text or image tokens.

The challenge lies in the conflicting objectives of the two tasks.
Comprehension pursues a many-to-one mapping that abstracts visual details (\eg, many photos of corgi dogs result in recognition of ``corgi'').
On the other hand, generation finds a one-to-one mapping that preserves visual details (\eg, the sketched image specific to the query image).
To bridge these conflicting objectives, the most straightforward way is to combine LLMs and diffusion models (DMs), which excel in comprehension and generation respectively. 
There are two main integration strategies. One approach is cascading~\cite{liu2023llava}, where the LLM understands user queries and plans instructions, which are then passed to the DM for generation. The other approach involves quantizing~\cite{esser2021taming, zhu2024scaling, lee2022autoregressive, yu2021vector, zheng2022movq} or mapping~\cite{dong2023dreamllm} the continuous visual modalities into discrete tokens for LLM processing, followed by decoding these tokens into images using the DM.
Recently, Transfusion~\cite{zhou2024transfusion} aims to combine the LLM and DM into a single model, creating a unified framework for both understanding and generating multimodal content.

\begin{figure}[t]
\centering
\includegraphics[width=0.48\textwidth]{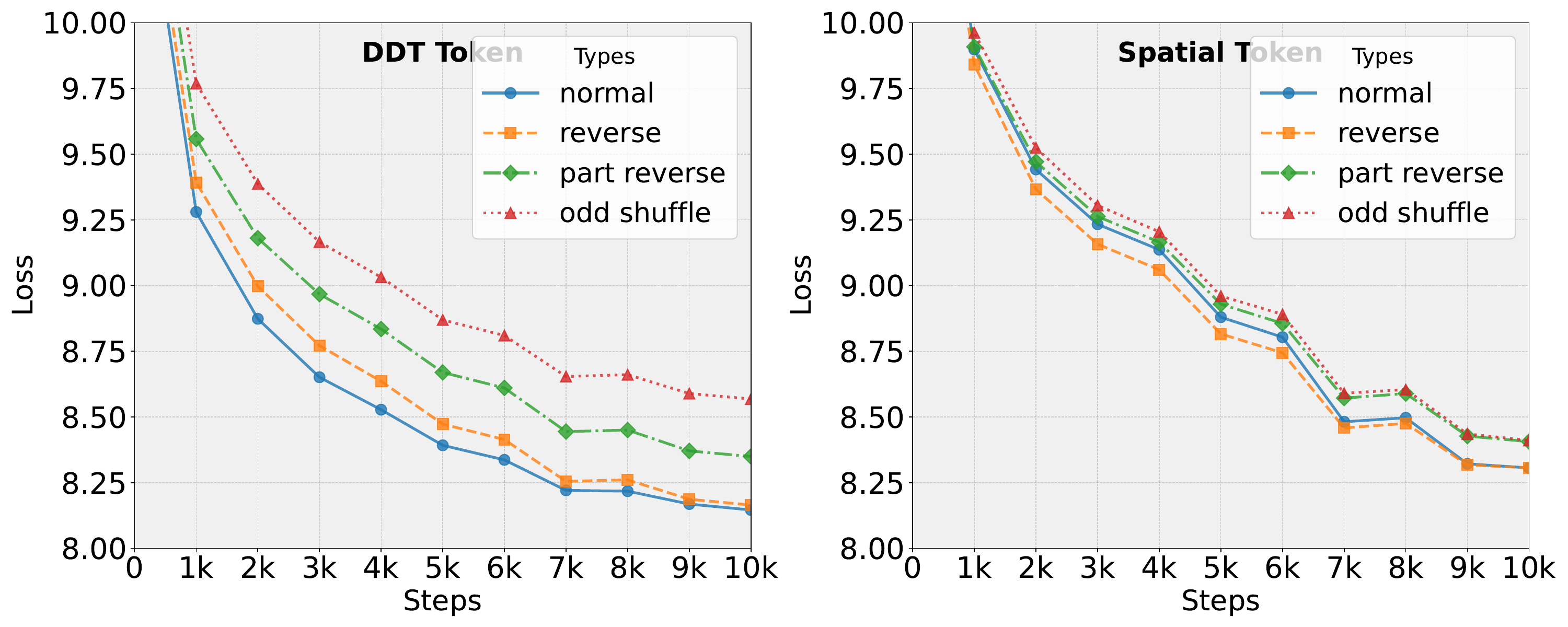}
\vspace{-2.0em}
\caption{Auto-regressive training curves of diffusion timestep tokens \textit{(left)} and spatial tokens \textit{(right)} under different degrees of sequence perturbation.
}
\vspace{-2em}
\label{fig:intro}
\end{figure}

Notably, all existing methods are based on spatial visual tokens, which are extracted from image patches and arranged in a spatial order (\eg, row by row in raster scan order).
However, following the evaluation in~\cite{kallini2024mission}, we find that spatial token sequences lack the traits of human language, and essentially form an \emph{impossible language}~\cite{kallini2024mission} for LLM to master.
Specifically, the performance of LLMs on language tokens should degrade rapidly when their order is disrupted.
However, when we train an LLM to model spatial token sequences in Figure~\ref{fig:intro} right, the order of tokens has minimal impact on the convergence of the LLM, deviating from phenomenons observed in human language.
This finding is perhaps non-surprising, as spatial tokens lack the recursive structure that characterizes languages. For example, one can extend ``a man'' by recursively embedding phrases to get ``a man [[walking] in the park]''.

To address the limitation of spatial tokens, we propose to learn tokens with recursive structure by Discrete Diffusion Timestep (DDT) tokenization.
Specifically, the forward process of a Diffusion Model (DM) incrementally adds Gaussian noise to
images at each time-step $t$, which collapses different images into similar ones by gradually losing the visual attributes that differentiate them.
We aim to learn an \emph{expanding} sequence of discrete tokens to make up for the \emph{incremental} attribute loss as $t$ increases.
For example, given a noise-free image $\mathbf{x}_0$, we obtain its $t$-specific token sequence $f_t(\mathbf{x}_0)$ by an encoder $f$, with an initial condition $f_0(\mathbf{x}_0)=\emptyset$ (\ie, no attribute loss).
At $t$, the noised image is given by $\mathbf{x}_t$. We train the DM to reconstruct $\mathbf{x}_0$ given $\mathbf{x}_t$ and $f_t(\mathbf{x}_0)$, so that the tokens compensate for the attribute loss in $\mathbf{x}_t$.
As $t$ increases, we learn a \emph{recursive} sequence given by $f_{t+1}(\mathbf{x}_0)=\left(f_t(\mathbf{x}_0\right), \textrm{V}_{t+1})$, where the appended visual token $\textrm{V}_{t+1}$ accounts for the increased attribute loss in $\mathbf{x}_{t+1}$ over $\mathbf{x}_t$.
Until at $t=T$, $x_T$ becomes pure noise where all attributes are lost. We output $f_T(\mathbf{x}_0)$ as the final token sequence for the image $\mathbf{x}_0$, which encodes all its visual attributes.
We do the same language verification on our DDT tokens in Figure~\ref{fig:intro} left. With our recursive tokens, we observe  findings similar to those on languages.

Using a DDT tokenizer trained on ImageNet~\cite{deng2009imagenet}, we build an MLLM that simultaneously unseals the potential of LLM in language modeling and DM in image generation.
First, we transform images into DDT tokens. Then we train an MLLM with a unified next-token-prediction objective on a vast corpus of image-text pairs (see Section~\ref{sec:method}), where the model essentially learns to translate between human language and the visual language using DDT tokens.
Finally in inference, when MLLM outputs visual tokens, they are mapped to images using our trained DM as decoder.
This leads to a unified framework for comprehension and generation.
Our main contributions are three-fold:
\begin{itemize}
    \item We propose DDT to learn discrete, recursive visual tokens by recovering visual attributes lost in the diffusion process that can subsequently drive an MLLM.
    \item We propose a novel way to integrate LLMs and diffusion models with our DDT tokens, producing an effective MLLM for both visual comprehension and generation.
    \item Despite using a tokenizer trained on just ImageNet, we achieve a new SOTA for multimodal comprehension and generation (\textit{i.e.,} image generation, image editing, visual comprehension), surpassing MLLMs that use tokenizers trained on large-scale data (\eg, Laion~\cite{schuhmann2022laion} and Coyo~\cite{kakaobrain2022coyo-700m}).
\end{itemize}

\section{Related Work}

Research on achieving unified understanding and generation in multimodal models has primarily focused on two main strategies: cascading architectures and tokenization-based methods. 
Cascading architectures~\cite{ge2023making, sun2024generative, jin2024unified, ge2024seed} integrate separate modality-specific encoders and decoders, each pre-trained independently, and then fuse their representations through projection layers to create combined models for multimodal tasks. 
Notable examples include models such as EMU2~\cite{sun2024generative}, which uses pre-trained language models~\cite{touvron2023llama} augmented with EVA-02-CLIP-E-plus~\cite{sun2023eva} for comprehension tasks, and cascade an SDXL-initialized~\cite{podell2023sdxl} diffusion model for visual generation tasks. 
In contrast, tokenization-based methods~\cite{team2024chameleon, pan2024auto, wu2024janus, wu2024vila, liu2024lumina, zhuang2025vargpt, chen2025janus, wang2024emu3} aim to create a unified framework by converting visual and textual inputs into a discrete space, and then jointly training a single transformer based solely on next-token prediction, thereby eliminating the need for diffusion or compositional approaches entirely.
Moreover, recent advances, such as TransFusion~\cite{zhou2024transfusion} and Show-o~\cite{xie2024show}, explore a blend of diffusion and autoregressive models within a single transformer for enhanced performance.
However, although these methods provide a step towards unification, most of them focus on spatial tokens for vision, which are extracted from image patches and arranged in a spatial order (\eg, row by row in raster scan order). These spatial tokens lack the traits of human language, resulting in an inability to seamlessly integrate with human natural language within an MLLM  to mutually enhance each other. 
Consequently, existing MLLMs still lag behind specialized architectures like SDXL~\cite{podell2023sdxl} in visual generation tasks and LLaVA-1.6~\cite{liu2024llavanext} in visual comprehension tasks.
These indicate the need for further exploration into more holistic tokenization methods that go beyond spatial representations to achieve more comprehensive multimodal comprehension and generation.
\begin{figure*}[t]
\centering
\includegraphics[width=\textwidth]{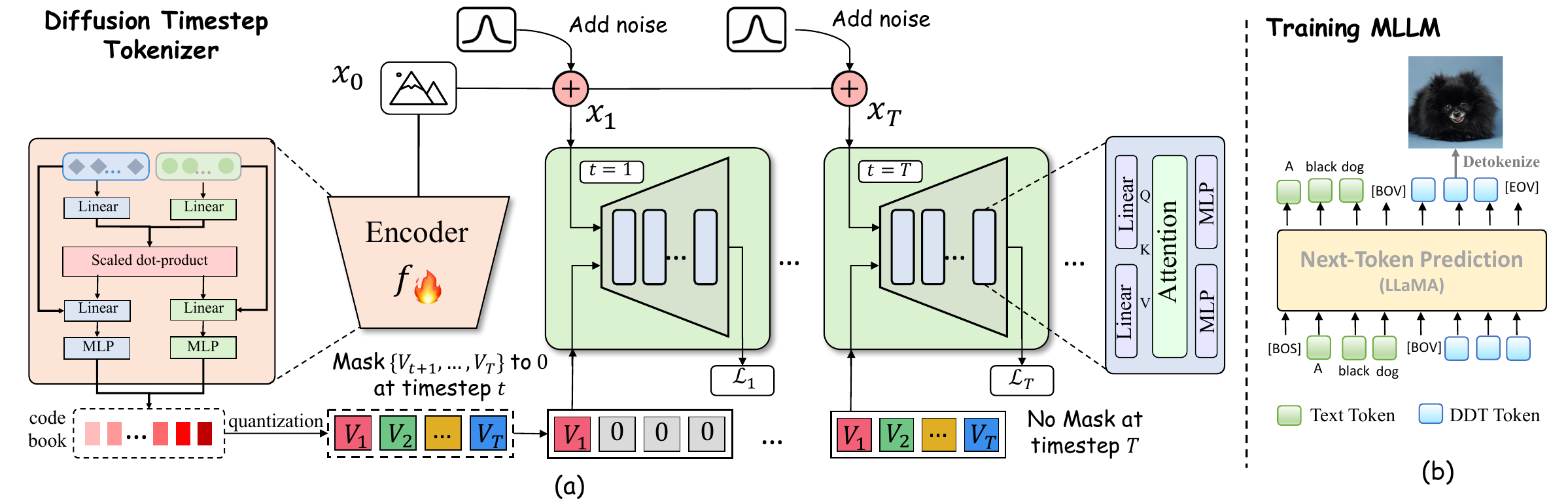}
\vspace{-2em}
\caption{The overview of our methods. \textit{\textbf{(a):}} The architecture of diffusion timestep tokenizer encodes an image to a recursive sequence of discrete tokens. \textit{\textbf{(b):}} An MLLM architecture that unifies comprehension and generation based on next token prediction. }
\vspace{-1em}
\label{fig:overview}
\end{figure*}

\section{Diffusion Timestep Tokenizer}
\label{sec:method}

Our goal is to train an image tokenizer that encodes an image to a recursive sequence of discrete tokens---to effectively train an MLLM---and decode the tokens back to the image---to render the image tokens predicted by the MLLM during inference.
The model consists of 1) an \textbf{encoder} $f$ that maps a noise-free image $\mathbf{x}_0$ to a sequence of continuous features $(\hat{\mathrm{V}}_1,\ldots,\hat{\mathrm{V}}_T)$; 2) a \textbf{quantizer} that assigns each continuous feature $\hat{\mathrm{V}}_i$ to its token embedding $\mathrm{V}_i \in \mathcal{C}$ in a fixed size dictionary $\mathcal{C}$; 3) a Diffusion Model (DM) $d$ as a \textbf{decoder} that reconstructs $\hat{\mathbf{x}}_0 = d\left( \mathbf{x}_t, t, (\mathrm{V}_1,\ldots,\mathrm{V}_t) \right)$, where $\mathbf{x}_t$ is a sampled noised image at timestep $t$ and the tuple denotes the first $t$ tokens from the quantizer. The model is trained end-to-end with (primarily) a reconstruction loss. 
And then we provide details on each part below.

\noindent\textbf{Encoder}. As shown in Figure~\ref{fig:overview}(a) left, we use a transformer-based model with learnable query tokens as our encoder. The input to the encoder is a noise-free image (patchified and flattened into image tokens similar to ViT~\cite{dosovitskiy2020image}) and the set of $T$ learnable query tokens. Note that the input query tokens have the same size $T$ as the desired output tokens. We first apply 1D and 2D sinusoidial position embedding on the query and image tokens, respectively. Following SD3~\cite{esser2024scaling}, we then use two independent transformers to process the query and image tokens, and join them for the attention operation. Finally, only the transformed query tokens are kept as the output $(\hat{\mathrm{V}}_1,\ldots,\hat{\mathrm{V}}_T)$, where $\hat{\mathrm{V}}_i \in \mathbb{R}^n$ is an $n$-dimensional feature vector.

\noindent\textbf{Quantizer}. We feed the encoder output to a standard Vector Quantization (VQ) module~\cite{van2017neural}. Specifically, the VQ has a fixed size dictionary $\mathcal{C}$ ($|\mathcal{C}|=65536$ in our experiments), where each entry is a vector of dimension $m$. We first use a linear layer to project the encoder output from $\mathbb{R}^n$ to $\mathbb{R}^m$, and then find $\mathrm{V}_i$ in $\mathcal{C}$ with the largest cosine similarity with each $\hat{\mathrm{V}}_i$. Note that we use an EMA-variant of VQ to improve training stability, \ie, updating $\mathcal{C}$ by gradually moving each entry towards the cluster center of its matched encoder outputs. To improve codebook usage, we use two tricks: 1) We use a small $m=16$ based on the empirical results in~\cite{yu2021vector}; 2) Inspired from~\cite{zeghidour2021soundstream}, we monitor dead entries in $\mathcal{C}$ at each training step (\ie, rarely matched with any $\mathrm{V}_i$) and setting them to random $\mathrm{V}_i$ in a training batch.

\noindent\textbf{Decoder}. Our decoder is a DM. As shown in Figure~\ref{fig:overview}(a) right, we use the same MMDiT~\cite{esser2024scaling} architecture for our decoder $d$ with slight modifications: 1) Instead of inputting text tokens, we input a sequence of quantized tokens; 2) We use a linear layer to project the $m$-dimensional vector to the latent dimension of the MMDiT; 3) We do not use pooled token embedding as in SD3. Given a timestep $t$, a sampled noisy image $\mathbf{x}_t$ from $\mathbf{x}_0$, a sequence of tokens from the quantizer, $d$ is trained to output a reconstructed $\mathbf{x}_0$ as shown below.

\noindent\textbf{Training}. 
Recall that we aim to learn a recursive sequence in the form $f_{t}(\mathbf{x}_0)=\left(f_{t-1}(\mathbf{x}_0\right), \textrm{V}_{t})$ at each $t$, such that $f_{t}(\mathbf{x}_0)$ makes up for the attribute loss in the noisy image $\mathbf{x}_t$. Hence we use an expanding set of tokens $({\mathrm{V}}_1,\ldots,{\mathrm{V}}_t)$ as the input to the decoder $d$, and train everything end-to-end with the reconstruction loss:
\begin{equation}
\small
    \mathcal{L} = \mathop{\mathbb{E}}_{t, \mathbf{x}_0, \epsilon} \; \underbrace{\left[ \lVert  d \left( t \epsilon + (1-t) \mathbf{x}_0 , t, ({\mathrm{V}}_1,\ldots,{\mathrm{V}}_t)\right) - \mathbf{x}_0 \rVert^2 \right]}_\text{$\mathcal{L}_t$},
    \label{eq:6}
\end{equation}
where we follow Rectified Flow~\cite{liu2022flow} to sample the noisy $\mathbf{x}_t = t \epsilon + (1-t) \mathbf{x}_0$ with the Gaussian noise $\epsilon$. We also use a standard commitment loss $\sum_{i=1}^T \lVert \hat{\mathrm{V}}_i - \mathrm{V}_i \rVert^2$ to regularize the encoder output (continuous vector) to be similar with the matched quantized vector.

\begin{table*}[ht]
    \centering
    \caption{ \label{tab:t2i} Comparison with state-of-the-art models on GenEval, T2I CompBench and DrawBench on zero-shot text-to-image generation. The best results
among MLLM-based Generalists are in \textbf{bold fonts}. And the best results among all methods are \underline{underline}. }
    \vspace{-1em}
    \resizebox{\linewidth}{!}{
        \begin{tabular}{l|ccccccc|cccccc|c}
            \toprule
                    & \multicolumn{7}{c}{\textbf{GenEval}}  &  \multicolumn{6}{c}{\textbf{T2I-CompBench}} &  \multicolumn{1}{c}{\textbf{DrawBench}}   \\
            Method  & Overall$\uparrow$  & SingObj$\uparrow$ &  TwoObj$\uparrow$ & Counting$\uparrow$  & Colors$\uparrow$ & Position$\uparrow$ & ColorAttri$\uparrow$ &  Color$\uparrow$  & Shape$\uparrow$ & Texture$\uparrow$  & Spatial$\uparrow$ & NonSpatial$\uparrow$ & Complex$\uparrow$ & Clip-T$\uparrow$\\
            \hline
            \multicolumn{15}{c}{\textit{Diffusion-based T2I Specialists}} \\
            \hline
            DALL-E 2  ~\cite{ramesh2022hierarchical} & 0.52 & 0.94 & 0.66 & 0.49 & 0.77 & 0.10 & 0.19   & 0.575 & 0.546 & 0.637 & 0.128 & 0.304 & 0.370 & -\\
            SDv1.5~\cite{rombach2022high}            & 0.43 & 0.97 & 0.38 & 0.35 & 0.76 & 0.04 & 0.06 & 0.373 & 0.365 & 0.422 & 0.131 & - & -  & -\\
            SDv2.1~\cite{rombach2022high}            & 0.50 & 0.98 & 0.51 & 0.44 & 0.85 & 0.07 & 0.17    & 0.569 & 0.450 &  0.498 & 0.174 & - & -  & -\\
            SDXL~\cite{podell2023sdxl}               & 0.55 & 0.98 & 0.74 & 0.39 & 0.85 & 0.15 & 0.23    & 0.637 & 0.541 & 0.564 & 0.203 & 0.311 & 0.409 & -\\
            PixArt-alpha~\cite{chen2023pixartalphafasttrainingdiffusion}       & 0.48 & 0.98 & 0.50 & 0.44 & 0.80 & 0.08 & 0.07    & 0.689  & 0.558 & 0.704  & \underline{0.208} & \underline{0.318} & \underline{0.412} & -\\
            DALL-E 3 ~\cite{betker2023improving}   & 0.67  & 0.96 & 0.87 & 0.47 & 0.83 & \underline{0.43} & 0.45    & \underline{0.811} & \underline{0.675} & \underline{0.807}   & - & - & - & -\\ 
            SD3~\cite{esser2024scaling} &  \underline{0.74}      & \underline{0.99} & \underline{0.94} & \underline{0.72} & \underline{0.89} & 0.33 & \underline{0.60}  & - & - & -  & - & - & - & -\\

            \hline
            \multicolumn{15}{c}{\textit{MLLM-based Generalist}} \\
            \hline
            SEED-LLaMA~\cite{ge2023making} & 0.33 & 0.89 & 0.17 & 0.25 & 0.59 & 0.02 & 0.03  & 0.176 & 0.184 & 0.202 & 0.068 & 0.289 & 0.227 & 0.220\\
            LaVIT~\cite{jin2024unified} & 0.51 & 0.93 & 0.70 & 0.43 & 0.67 & 0.17 & 0.18  & 0.439 & 0.457 & 0.575 & 0.151 & 0.311 & 0.320 & 0.259\\
            Emu2-Gen~\cite{sun2024generative} & 0.35 & 0.93 & 0.28 & 0.14 & 0.70 & 0.05 & 0.02  & 0.338 & 0.359 & 0.400 & 0.099 & 0.299 & 0.285 & 0.235\\
            SEED-X~\cite{ge2024seed} &  0.49&0.96 & 0.57 & 0.29 & 0.82 & 0.14 & 0.15  & 0.657 & 0.492 & 0.603 &  0.172 & 0.306 & \textbf{0.361} & 0.265 \\
            VILA-U~\cite{wu2024vila} & 0.40 & 0.88 & 0.42 & 0.25 & 0.69 & 0.08 & 0.09 & 0.568 & 0.433 & 0.501 & 0.154 & 0.307 & 0.334 & 0.262\\
            Lumina-mGPT~\cite{liu2024lumina} & 0.52 & 0.99 & 0.70 & 0.34 & 0.85 & 0.08 & 0.17 & 0.556 & 0.465 & 0.588 & 0.176 & 0.312 & 0.343 &\textbf{0.267} \\
            Transfusion~\cite{zhou2024transfusion}   & 0.63 & - & - & - & - & - & - & - & - & - & - & - & -  & -\\
            Emu3~\cite{wang2024emu3}  & 0.54 & 0.98 & \textbf{0.71} & 0.34 & 0.81 & 0.17 & 0.21  & 0.611 & 0.473 & 0.618 & 0.137 & 0.300 & 0.345  & 0.265\\ \hline
            \textbf{Ours} & \textbf{0.66} & \textbf{0.99} & 0.64 & \textbf{0.56} & \textbf{0.87} & \textbf{0.39} & \textbf{0.48} & \textbf{0.728} & \textbf{0.514} & \textbf{0.642} & \textbf{0.176} & \textbf{0.314} & 0.345 & \textbf{0.267}\\

            \bottomrule
        \end{tabular}
    }
    \vspace{-1.2em}
\end{table*}

\section{DDT-LLaMA}
\label{MLLM}
The process begins by converting images into a 1D recursive discrete sequence like a foreign language that LLM can read.
This transformation allows us to pursue a unified next-token-prediction training objective across a vast array of image-text data. To achieve this, visual tokens are concatenated with text tokens to form a multi-modal sequence. Special tokens, \texttt{[BOV]} and \texttt{[EOV]}, are strategically placed at the beginning and end of the visual sequence to distinguish between modalities.
We Initialize the MLLM from a pre-trained LLM, specifically the LLaMA3-8B~\cite{dubey2024llama} model, with an expansion of its vocabulary by 65,536 visual codes. 
Then MLLM adopts the general Language Modeling (LM) objective to directly maximize the likelihood of each multi-modal sequence in an auto-regressive manner:
\begin{equation}
    p(y) = \sum_{y \in \mathcal{D}} \sum_{i=1}^{S} \log P_\theta(y_i | y_{< i}).
\end{equation}
where $y$ is the text or visual token with $S$ as the sequence length. Since both image and text are represented as discrete token IDs, we can use the cross-entropy to supervise the token prediction at each position for both modalities with a shared prediction head. The training regimen is composed of two stages, during which all parameters of the MLLM are meticulously fine-tuned.

\paragraph{Stage1: Pre-training.} In the first stage, we aim to enhance the alignment between DDT tokens and text tokens, transitioning from LLM to MLLM. 
Our pre-training dataset, sourced from Laion~\cite{schuhmann2022laion} and Coyo~\cite{kakaobrain2022coyo-700m}, consists of 200M text-to-image pairs. Each pair is structured in the format ``\texttt{[BOS] <caption text> [BOV] <DDT tokens> [EOV] [EOS]}'', where \texttt{[BOS]} and \texttt{[EOS]} are special tokens from the text tokenizer.
Moreover, to prevent the forgetting of textual capabilities, we also incorporate pure text data at a fusion ratio of 10\%. The pre-training is conducted on 512 Ascend 910B NPUs, lasting for nearly two weaks.

\paragraph{Stage2: Instruction Tuning.} 
In the second stage, we perform multimodal instruction tuning to align our MLLM with human instructions through supervised fine-tuning on public image-text datasets. 
To enhance the MLLM’s comprehension and generation capabilities in complex scenarios, we employ supervised training data from visual comprehension and generation tasks. We use the instruction format as ``\texttt{USER: <Instructions> ASSISTANT: <Answers>}'' and only the content of \texttt{<Answer>} is accounted for loss. 
See Appendix A and B for more details.

\paragraph{Inference.} After DDT-LLaMA predicts \texttt{[BOV]} token, it consecutively performs $T$ next-token-predictions to generate a set of discrete DDT tokens $\{V_1,...,V_T\}$. We then feed these tokens into the decoder to guide the denoising of random noise into an image over $T$ timesteps through a DDPM process. Similar to the training procedure, at timestep $t$ during inference, we provide only the first $t$ tokens as the guidance, while the remaining $(T-t)$ tokens are masked as $0$.

\section{Experiment}

\begin{figure*}[t]
\centering
\vspace{-1em}
\includegraphics[width=0.95\textwidth]{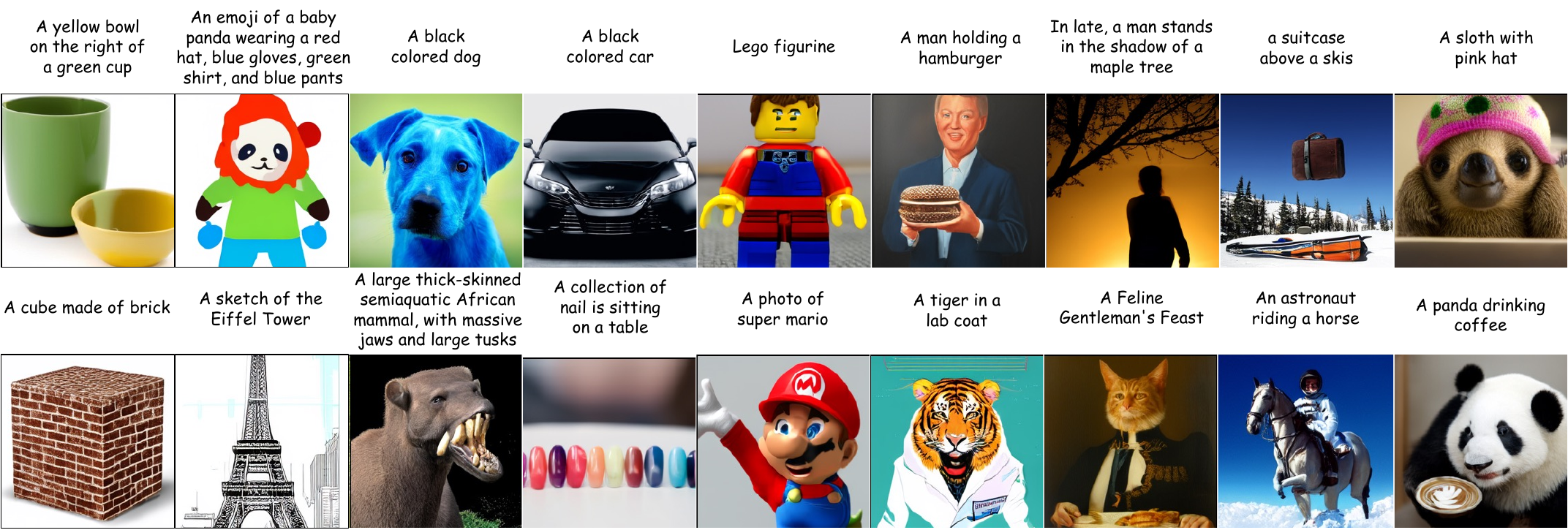}
\vspace{-1em}
\caption{Qualitative results of DDT-LLaMA text-to-image generation.}
\vspace{-1.3em}
\label{fig:t2i}
\end{figure*}

\begin{table}[t]
    \centering
    \caption{\label{tab:edit} Zero-shot image editing results.}
    \vspace{-1em}
    \begin{adjustbox}{width=0.48\textwidth}
    \begin{tabular}{lcccccccc}
    \toprule
    ~  & \multicolumn{2}{c}{\textbf{EVR}} & ~ & \multicolumn{2}{c}{\textbf{MA5K}} & ~ & \multicolumn{2}{c}{\textbf{MagicBrush}} \\
    \cmidrule{2-3} \cmidrule{5-6} \cmidrule{8-9} Method  & L1$\downarrow$ & CVS$\uparrow$  & ~ & L1$\downarrow$ & LPIPS$\downarrow$  & ~ & L1$\downarrow$ & CVS$\uparrow$  \\ \hline
    \multicolumn{9}{c}{\textit{Image editing Specialist}} \\ \hline
    InsPix2Pix~\cite{brooks2023instructpix2pix}      & 18.9     &  81.4    & ~     &  17.6   &35.9  & ~     & 10.1    &85.2 \\
    MGIE~\cite{fu2023guiding}     &    16.3 &  81.7    & ~     & \underline{13.3}     &  29.8 & ~     & 8.2     &  91.1\\ 
    UltraEdit~\cite{zhao2024ultraedit}    &    \underline{14.4} &  81.6    & ~     & 16.5     &  35.2 & ~     & \underline{6.6}     &  88.4\\ \hline
    \multicolumn{9}{c}{\textit{MLLM-based Generalist}} \\ \hline
    GILL~\cite{koh2024generating} & 31.8 & 65.0 & ~ & 27.4 & 44.3  & ~& 28.3 & 75.2 \\
    Emu2-Gen~\cite{sun2024generative}     & 22.8     & 80.3     & ~     & 20.5     &36.4  & ~     & 19.9     & 85.7 \\
    SEED-LLaMA~\cite{ge2023making}        & 28.4     & 72.3     & ~     & 24.6     &39.0  & ~     & 24.5     & 80.9 \\
    LaVIT~\cite{jin2024unified}   & 26.8     & 73.8     & ~     & 25.1     & 36.9  & ~     & 25.3     & 81.1 \\
    SEED-X-Edit~\cite{ge2024seed}   & 16.7     & 81.9     & ~     & 14.6     & 32.7  & ~     & 18.4     & 86.1 \\ \hline
    \textbf{Ours}        & \textbf{14.5}    & \underline{\textbf{85.5}}     & ~     & \textbf{14.2}    & \underline{\textbf{29.6}}  & ~     & \textbf{7.1}     & \underline{\textbf{92.4}} \\ 
   
    \bottomrule
    \end{tabular}
    \end{adjustbox}
    \vspace{-1em}
    
    % \vspace{-0.5em}
    \vspace{-1em}
\end{table}

\begin{figure*}[t]
\centering
\includegraphics[width=\textwidth]{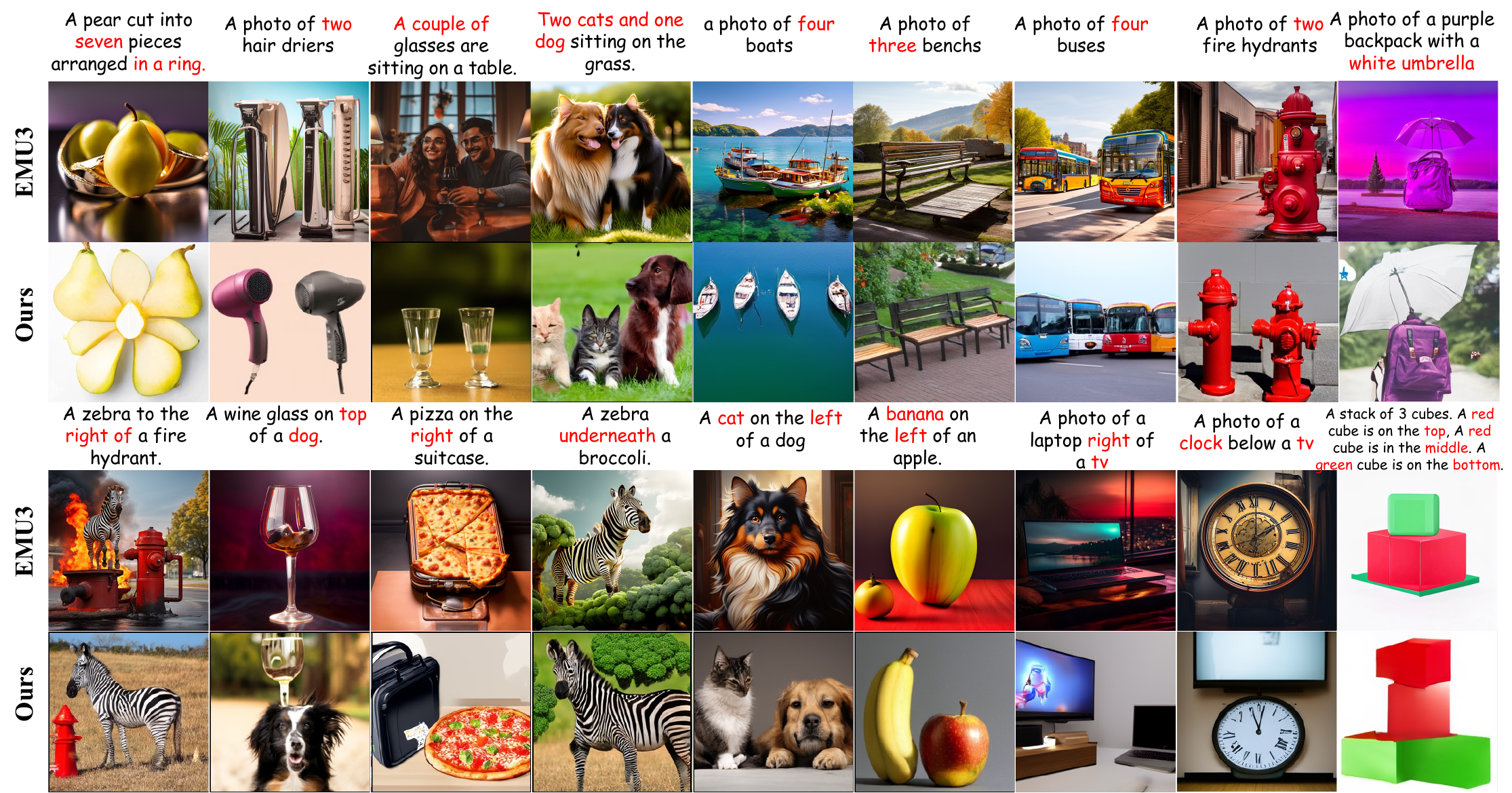}
\vspace{-1.5em}
\caption{Qualitative comparison with EMU3  on T2I generation. DDT-LLaMA better responses to prompts related to counting or position.}
\vspace{-1em}
\label{fig:emu3}
\end{figure*}

\subsection{Text-to-Image Generation}
% \subsubsection{Automated Metric Evaluation}
\paragraph{Automated Metric Evaluation. }
To evaluate our model's capabilities in zero-shot image synthesis, we first conduct an automated metric evaluation on 3 popular text-to-image benchmarks: GenEval~\cite{ghosh2024geneval}, T2I-CompBench~\cite{huang2023t2icompbench}, and DrawBench~\cite{saharia2022photorealistic}. 
The comparison results of DDT-LLaMA against Diffusion-based T2I Specialists, and MLLM-based generalists are shown in Table~\ref{tab:t2i}.
In most settings, our method surpasses other MLLMs and is even competitive with several diffusion-based models.
For example, in the overall GenEval score, DDT-LLaMA significantly outperforms SEED-X~\cite{ge2024seed}, Emu3~\cite{wang2024emu3}, and the specialist method, SDXL~\cite{podell2023sdxl}. 
This highlights that our model facilitates better vision-text alignment.

Furthermore, through detailed analysis of the results, we find that compared to other MLLMs, DDT-LLaMA achieves higher scores in tasks related to color, counting, and position. 
It excels in understanding object attributes such as color, quantity, and the spatial relationships between objects.  
However, it underperforms in tasks of 'TwoObj', which simply requires generating two objects with no additional constraints. This is largely due to the fact that \textit{the image-text captions in our training data rarely include this straightforward prompt format}.

\begin{figure*}[t]
\centering
\includegraphics[width=\textwidth]{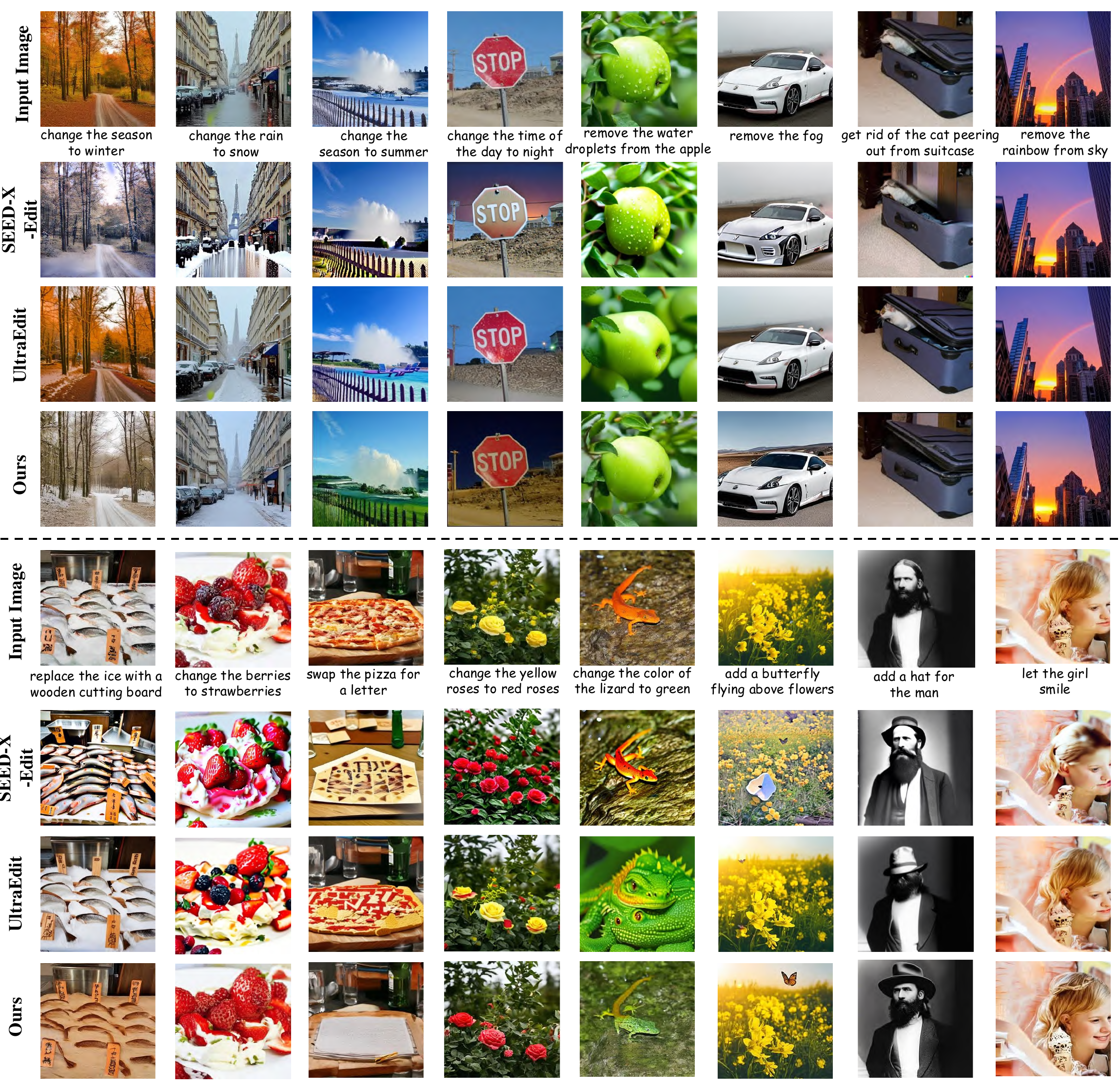}
\vspace{-2em}
\caption{Qualitative comparison on image editing. }
\vspace{-1em}
\label{fig:edit}
\end{figure*}

\paragraph{Qualitative Examples. }

In Figure~\ref{fig:t2i}, we present some qualitative examples of DDT-LLaMA on image synthesis tasks. 
It can be observed that DDT-LLaMA effectively follows various types of instructions, including complex ones such as generating surreal images and multi-condition combined prompts, to generate semantically consistent images.
Of course, we acknowledge that there is room for improvement in the aesthetic quality of the images. This limitation stems from the fact that our current tokenizer was only trained on ImageNet with a resolution of 256x256. So its performance in reconstructing open-domain images is inferior to that of baseline methods (\textit{e.g.}, the EMU3 tokenizer MOVQ), which restricts our capability in T2I generation. More qualitative examples are in Appendix C.

Moreover, Figure~\ref{fig:emu3} presents a direct comparison between DDT-LLaMA and Emu3 on prompts involving counting, color, and position. Though Emu3 generates images with higher aesthetic quality, it struggles to accurately respond to such prompts. In contrast, DDT-LLaMA generates images that \textbf{ better align with the object attributes and the spatial relationships specified in the prompt.}

\subsection{Image Editing}
\paragraph{Automated Metric Evaluation. }
We evaluate our model in zero-shot instruction-based image editing across three datasets: EVR~\cite{tan2019expressing}, MA5k~\cite{shi2021learning}, and MagicBrush~\cite{zhang2023magicbrush}. 
Following \cite{fu2023guiding}, for EVR and MagicBrush, we treat L1 and CVS (Clip-Visual-Similarity)  as the evaluation metrics. For MA5K, we utilize L1 and LPIPS~\cite{zhang2018unreasonable}) for evaluation.
The results are shown in Table~\ref{tab:edit}.

We find that DDT-LLaMA demonstrates significant superiority over existing MLLMs across all test datasets and metrics. It not only effectively comprehends the instruction to execute editing, but also excels at preserving image fidelity, which is rarely seen in prior work. 
Furthermore, when compared with specialist models, DDT-LLaMA still shows stronger performance. It consistently outperforms InsPix2Pix~\cite{brooks2023instructpix2pix} , also showing greater effectiveness than MGIE~\cite{fu2023guiding} and UltraEdit~\cite{zhao2024ultraedit} across most datasets.

\paragraph{Qualitative Examples. }
Figure~\ref{fig:edit} shows some qualitative image editing examples of DDT-LLaMA and baselines\cite{ge2024seed, zhao2024ultraedit}.
DDT-LLaMA supports a wide spectrum of editing operations including both local change (e.g., removal, replacement) and global change (change time, manipulation).
For each case, our output images not only resemble the source image to a high degree but also are coherent with the instruction, demonstrating that DDT-LLaMA achieves a great trade-off between fidelity and editability.

\begin{table*}[t]
    \centering
    % \vspace{-1em}
    \caption{\label{tab:comp} Comparison for multimodal comprehension on vision-language benchmarks.}
    \vspace{-1em}
    \begin{adjustbox}{width=\textwidth}
    \vspace{-1em}
    \begin{tabular}{l|cc|cc|cccc|ccc}
    \toprule
    \multirow{2}{*}{Models} & \multirow{2}{*}{Size} & \multirow{2}{*}{Image Gen} & \multicolumn{2}{c|}{Image caption} &  \multicolumn{4}{c|}{VQA} &  \multicolumn{3}{c}{MLLM Bench} \\
     &&                                       &NoCaps     &Flickr30K	 & VQA & GQA & 	OKVQA & VizWiz	& MME & SEEDBench & POPE  \\
    \hline
    \multicolumn{12}{c}{\textit{Specialized Visual Comprehension MLLMs}} \\ \hline
    InstructBlip~\cite{instructblip}                          & 7B    & \ding{55}    & 123.1  & 82.4  & -      & 49.2     & -      & 34.5   &1212.8     & 58.8   &-      \\
     QWenVL-Chat~\cite{Qwen-VL}                                   & 7B   & \ding{55}    & 120.2         & 81.0     & 78.2      &  57.5          &56.6      &38.9         &1487.57    &58.2      &-      \\
    LLaVA-1.5~\cite{liu2024improved}                                   & 7B   & \ding{55}    & 105.5         & -     & 78.5      &  62.0         &53.4      &50.0         &1510.7     &64.3      &85.9      \\
     mPLUG-Owl2~\cite{ye2024mplug}                               & 7B    & \ding{55}  & -         & 85.1     & 79.4      &  56.1          &57.7      &54.5         &1450.3     &57.8      &86.2      \\ 
     ShareGPT4V~\cite{chen2023sharegpt4v}                               & 7B    & \ding{55}  & -         & -     & 80.6      &  63.3         &-      &57.2         &1567.4     & \underline{69.7}      &-      \\ 
      LLaVA-1.6(HD)~\cite{liu2024llavanext}                               & 7B    & \ding{55}  & 88.3         & -     & 80.6      &  64.2           &44.3      & \underline{60.6}         &-     &64.7      &86.5      \\ 
    VILA~\cite{lin2023vila}                                   & 7B   & \ding{55}    & -         & -     & 79.9      &  62.3            &-      &57.9         &\underline{1533.0}     &61.1      &85.5      \\ \hline 
    \multicolumn{12}{c}{\textit{MLLMs for both Visual Comprehension and Generation}} \\ \hline
    Emu2-Chat~\cite{sun2024generative}                               & 37B    & - & 119.8        & 86.0         & \underline{\textbf{84.9}}     & 65.1          & 64.8      &54.9              &1315.7     &62.8      &-      \\    
    SEED-LLaMA~\cite{ge2023making}                              & 8B    &$\checkmark$& 90.4         & 66.7     & 66.2      &  34.8          & 45.9      &55.1                   &736.7 & 51.5 &-    \\
    VILA-U~\cite{wu2024vila}                              & 7B    &$\checkmark$& 95.4         & 70.3     & 75.3      &  58.3           & 59.6      & 35.3             &1336.2      &56.3   &83.9    \\
    LaVIT~\cite{jin2024unified}                                   & 7B    &$\checkmark$& 114.2         & 83.0     & 66.0      &  46.8           & 54.6      & 38.5         &997.9     &57.3      &82.2      \\
    Emu3~\cite{wang2024emu3}                                   & 8B    & - & 117.5         & 82.5     & 75.1     &  60.3          & 61.2      & 50.5         &-     & \textbf{68.2}      & 85.2      \\
    \hline
    \textbf{Ours}                           & 8B    &$\checkmark$ & \underline{\textbf{124.2}}         & \underline{\textbf{88.5}}     & 76.4      &  \underline{\textbf{66.3} }        & \underline{\textbf{65.9}}      &\textbf{57.3}         &\textbf{1378.9}     &67.9      & \underline{\textbf{86.9}}      \\
    \hline
    % \bottomrule
    \end{tabular}
    \end{adjustbox}
    \vspace{-1em}
    % \vspace{-1em}
\end{table*}

\subsection{Vision-Language Comprehension}
We also evaluate the vision-language comprehension capabilities across various public benchmarks~\cite{agrawal2019nocaps, plummer2015flickr30k, antol2015vqa, hudson2019gqa, marino2019ok, gurari2018vizwiz, fu2024mmecomprehensiveevaluationbenchmark, li2024seedbench, li2023evaluating}.
The primary results, detailed in Table~\ref{tab:comp}, compare two categories of methods: specialized visual comprehension MLLMs~\cite{instructblip, Qwen-VL, liu2023llava, ye2024mplug, chen2023sharegpt4v, liu2024llavanext, lin2023vila} and MLLMs designed for both visual comprehension and generation~\cite{sun2024generative,ge2024seed,wu2024vila,jin2024unified,wang2024emu3}. 
Notably, DDT-LLaMA is an encoder-free MLLM without pretrained encoders such as CLIP~\cite{radford2021learning}. Compared to other MLLMs, it significantly surpasses its counterparts across multiple benchmarks, even without relying on a specialized pretrained CLIP. It also outperforms many MLLMs that focus solely on comprehension tasks, highlighting the inherent capabilities and promising potential of DDT tokens in multimodal understanding.

\subsection{In-Depth Analysis}

\begin{table}[t]
    \centering
    \caption{\label{tab:fid} Reconstruction quality on ImageNet validation split and FID score comparison for autoregressive class-conditional synthesis on 256 $\times$ 256 ImageNet.}
    \vspace{-1em}
    \begin{adjustbox}{width=0.4\textwidth}
    \begin{tabular}{lcccccccc}
    \toprule
   Method & Reconstruction (PSNR) $\uparrow$ & Class-cond (FID) $\downarrow$\\ \midrule
    VQGAN~\cite{esser2021taming} & 18.9 & 15.8 \\
    VQGAN-LC~\cite{zhu2024scaling} & 20.8 & 15.4 \\
    RQ-Transformer~\cite{lee2022autoregressive} & - & 7.6 \\
    % Vit-VQGAN~\cite{yu2021vector} & - & 5.3 \\
    MoVQ~\cite{zheng2022movq} & 21.5 & 7.1 \\
    \hline
    \textbf{Ours} & \textbf{21.7} & \textbf{6.1} \\
   
    \bottomrule
    \end{tabular}
    \end{adjustbox}
    \vspace{-1.5em}

\end{table}

\subsubsection{Class-conditional Generation of DDT}
Our tokenizer is trained on 256×256 images from the ImageNet train split. The PSNR between the reconstructed and original images from the validation split reaches 21.7, outperforming spatial tokenizers trained on ImageNet (using a comparable number of tokens).

To directly compare the performance of DDT tokens with other spatial tokens for image generation, we also train an autoregressive GPT model for class-conditional image generation, following ~\cite{esser2021taming}. We compare our model with several previous SOTA spatial tokenizers~\cite{esser2021taming, lee2022autoregressive, zheng2022movq} for autoregressive modeling of class-conditional image synthesis on ImageNet. As shown in Table~\ref{tab:fid}, under the same autoregressive paradigm, our method achieves a superior FID score, demonstrating stronger performance than tokenizers that generate spatially-based visual tokens.

\begin{figure}[t]
\centering
\includegraphics[width=0.4\textwidth]{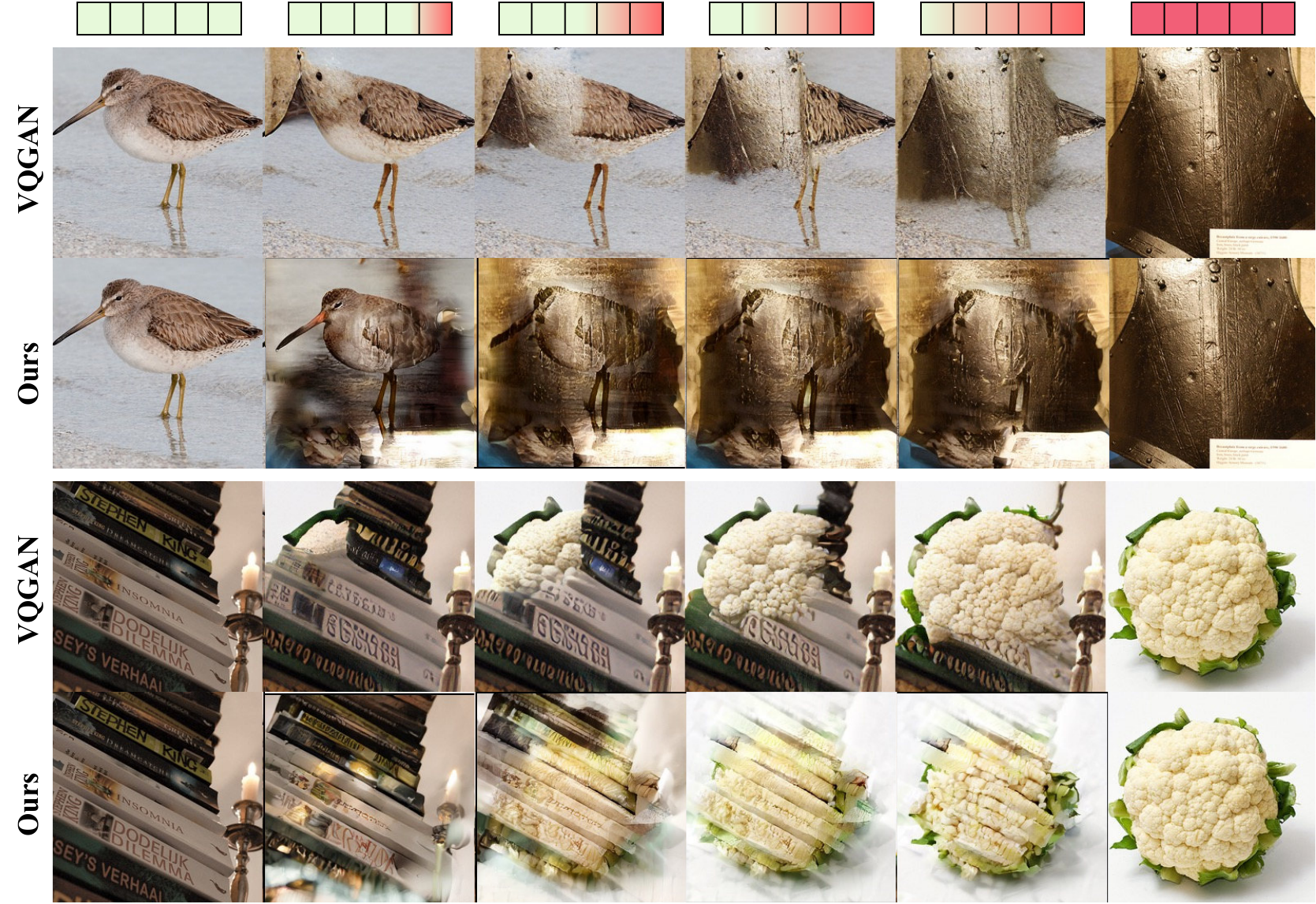}
\vspace{-1em}
\caption{Results of Counterfactual Interpolation with DDT tokens and VQGAN tokens. }
\vspace{-1.5em}
\label{fig:interpolation}
\end{figure}

\begin{figure*}[t]
\centering
\includegraphics[width=\textwidth]{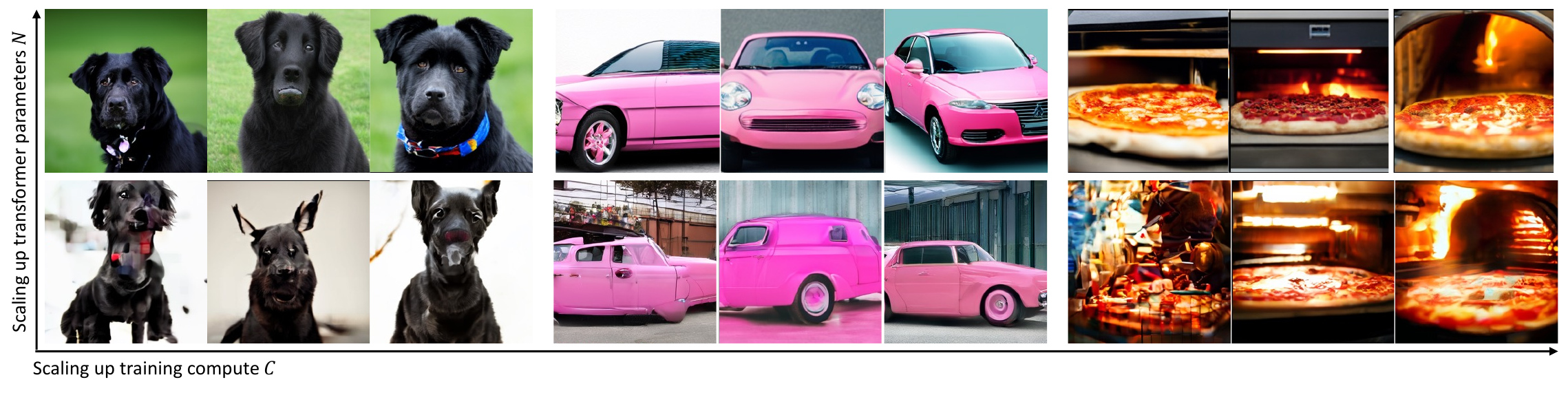}
\vspace{-2.5em}
\caption{Scaling MLLM size (2B, 8B) and training compute (50\%, 75\%, 100\% of total tokens) improves visual quality. }
\vspace{-1.0em}
\label{fig:scaling}
\end{figure*}

\subsubsection{DDT Tokens are Recursive}
\paragraph{Counterfactual Interpolation.} To directly show that DDT tokens are recursive and decouple visual attributes at various granularities, we first conduct counterfactual interpolation with both DDT tokens and VQGAN tokens. 
Specifically, for the token sequence derived from an image, we replace a subset of tokens with those from another image, while keeping the remaining tokens fixed, with the resulting sequence fed into the decoder for image generation.

As shown in Figure~\ref{fig:interpolation}, for VQGAN tokens, counterfactual interpolation actually performs CutMix~\cite{yun2019cutmix}, where regions from two images are concatenated. 
In contrast, DDT tokens, with their disentangled representation, ensure that only the attributes captured by the substituted tokens change in the generated counterfactuals, which allows for a seamless semantic blending of the two images.

\paragraph{Decoding Image with an expanding subset of DDT tokens.}
To further illustrate the recursive nature of DDT tokens, we only input a subset of the first $t$ timestep tokens that are autoregressively sampled by DDT-LLaMA, into the decoder for image synthesis ($t$ ranges from 1 to 480). As shown in  Figure~\ref{fig:mask}, with the number of tokens increasing, the image’s attributes are progressively recovered. 
Initially, the decoder reconstructs fine details, with contours and color information gradually completed until the full structure of the image is outlined.
This finding further demonstrates that DDT tokens are recursive, and DDT-LLaMA effectively disentangles visual attributes, progressively inferring visual features at varying granularities based on textual instructions. This behavior aligns with our expectations for the DDT token design.

\begin{figure}[t]
\centering
\vspace{-1em}
\includegraphics[width=0.4\textwidth]{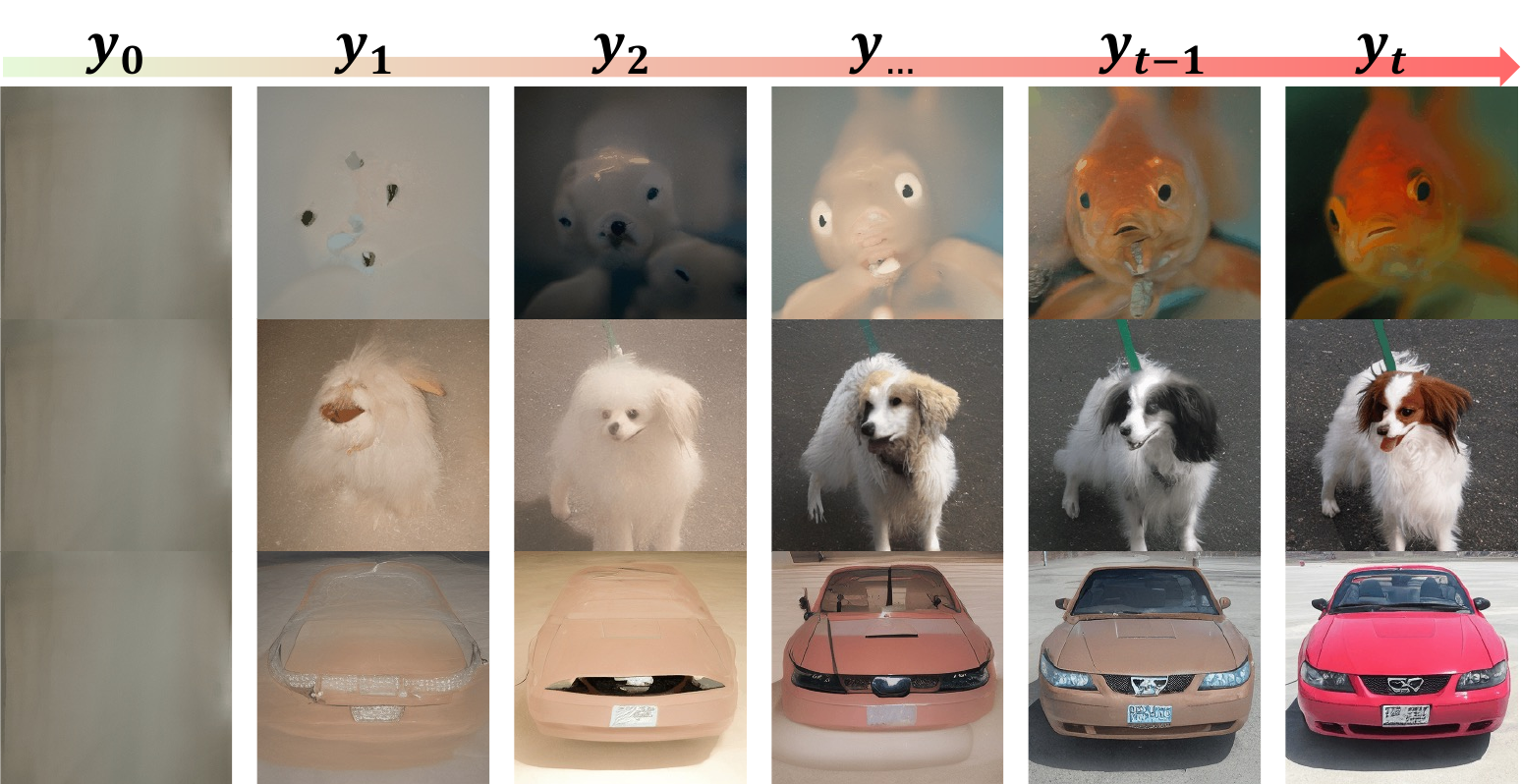}
\vspace{-1em}
\caption{Decoding images with an expanding subset of autoregressive-sampled DDT tokens. }
\vspace{-1.2em}
\label{fig:mask}
\end{figure}

\subsubsection{Perplexity of Text Generation}
To further demonstrate that our recursive DDT tokens can facilitate the synergism between comprehension and generation, we record the perplexity of text prediction during auto-regressive multimodal pre-training, with VQGAN tokens and DDT tokens, respectively.
As shown in Figure~\ref{fig:chart}(a), training with DDT tokens results in a decrease in perplexity for text generation, whereas training with VQGAN tokens yields the opposite conclusion.
This highlights that our DDT tokens are better suited for unified autoregressive modeling with text tokens, much like a foreign language that LLMs can read.

\subsubsection{A/B Test on Image Editing}
\label{sec:gemma}
\begin{figure}[t]
\centering
\vspace{-0.5em}
\includegraphics[width=0.46\textwidth]{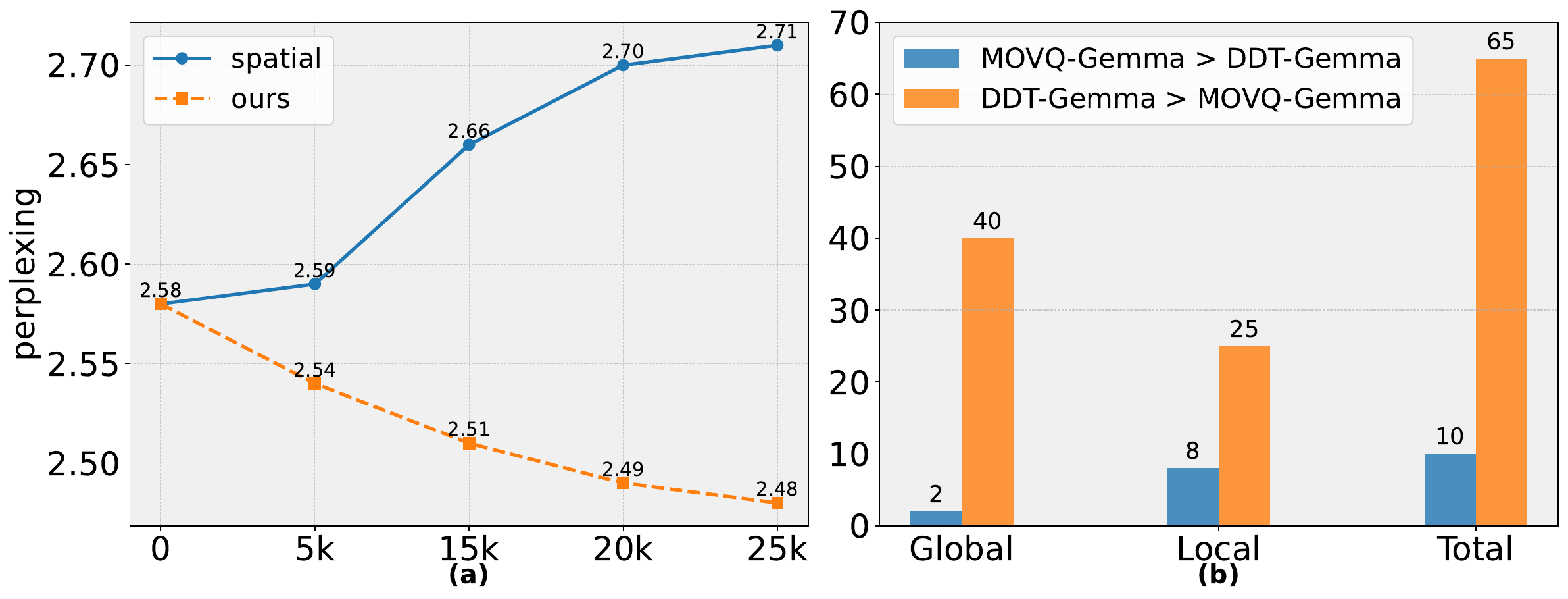}
\vspace{-1.1em}
\caption{(a) Perplexity of MLLM in text generation during image-generation training. (b) A/B test on image editing.}
\vspace{-1.2em}
\label{fig:chart}
\end{figure}

By disentangling visual attributes, DDT tokens hold an advantage over spatial tokens for tasks that require attribute modifying, such as image editing. 
To demonstrate this, we first use Gemma2-2B~\cite{team2024gemma2} as the backbone and leverage both DDT tokens and MoVQ tokens on the same 200M image-text dataset for pre-training, respectively. 
We then fine-tuned both models on the same image editing training set for consistent training steps. 
For evaluation, we manually collect a set of image editing test data and conduct an A/B test for human assessment.
As shown in Figure~\ref{fig:chart}(b), DDT-Gemma outperforms MoVQ-Gemma in 65 editing cases in total, showing a notable advantage in both global and local editing instructions. While MoVQ-Gemma surpasses DDT-Gemma in only 10 cases. It further underscores the superiority of DDT tokens.

\subsubsection{Scaling Laws of DDT-based MLLM}
We have also observed preliminary indications of scaling laws in our DDT-based MLLM. 
As mentioned in Sec~\ref{sec:gemma}, we also leverage Gemma2-2B for pre-training. And we further compare the performance of T2I generation with two model sizes (2B,8B) at three different training stages (50\%, 75\%, and 100\% of total training tokens), as shown in Figure~\ref{fig:scaling}. The observed improvements in visual quality align with scaling laws, which suggest that larger transformers trained on more extensive datasets can learn more detailed and fine-grained image distributions.

\section{Conclusion}
In this work, we propose Discrete Diffusion Timestep (DDT) tokenizer to learn discrete, recursive visual tokens, which recursively compensate for the progressive attribute loss in noisy images as timesteps increase.
With image decoding as DDT tokens, we train DDT-LLaMA on a vast corpus of image-text pairs for vision-language alignment.
Extensive experiments showcase the immense potential of DDT-LLaMA across various tasks, \textit{e.g.}, T2I generation, image editing, and vision-language understanding.
\textbf{\textit{
We are currently working on scaling up the training of our DDT tokenizer and the MLLM. 
In the near future, we will release a more powerful version of DDT-LLaMA, along with a more detailed technical report. Stay tuned!}}

% \clearpage
\section*{Acknowledgements}

This work was supported by the NSFC (62272411), the Key R\&D Projects in Zhejiang Province (No. 2024C01106, 2025C01030), the Zhejiang NSF (LRG25F020001). 
{
    \small
    \bibliographystyle{ieeenat_fullname}
    \bibliography{main}
}

\clearpage
\maketitlesupplementary

\appendix

\addtocontents{toc}{\protect\setcounter{tocdepth}{2}}
\paragraph{Overview}
In this supplementary material, we present the following content:
\vspace{-3em}
\renewcommand{\contentsname}{}
\tableofcontents
\appendix

\section{Implementation Details}
\subsection{Tokenizer}
\paragraph{Encoder.} 
In the encoder, we set $T=480$ (\ie, the number of query tokens). The dimensions of the query tokens and the noise-free image are $R^{480\times 256}$ and $R^{1024\times 256}$, respectively.
The encoder contains two independent transformers, each comprising $20$ layers with latent dimension of $256$. 
Following SD3~\cite{esser2024scaling}, despite the noise-free images and query tokens being input into separate transformers, we join the sequences of the two for the attention operation. This allows both representations to operate independently while considering the influence of the other. 
The encoder output retains only the transformed query tokens, serving as the image's latent representations.

\paragraph{Quantizer.} The quantizer is an EMA-variant of vector quantization. Following \cite{yu2021vector}, we leverage a linear projection from the encoder output to low-dimensional variable space for code index lookup (\textit{i.e.}, reduced from a $256$-d vector to a $16$-d vector per code). We also apply L2 normalization on the encoded latent features and codebook latent variables. Moreover, at each training step, we reset the dead entries in the codebook $\mathcal{C}$ (\ie, rarely matched with any tokens) to random tokens in the training batch. 

\paragraph{Decoder.}  we use the same MMDiT architecture proposed in SD3~\cite{esser2024scaling}  for our decoder with minor modifications.
Each transformer in the MMDiT comprises $24$ layers with latent dimension of $1536$. 
The sequence of quantized tokens replaces the text tokens as input, with a linear layer to project the $16$-dimensional quantized vector to the latent dimension (\textit{i.e.}, $1536$) of the MMDiT. Additionally, we also removed the pooled token embedding introduced in SD3.

Furthermore, the hyper-parameters of training the tokenizer are detailed in \textcolor{red}{\textbf{Table~\ref{tab:hyper}}}.

\begin{table}[t]
    \centering
    \caption{\label{tab:hyper}The detailed training hyper-parameters. ``MLLM-pt'' denotes the pretraining of DDT-LLaMA, ``MLLM-ft'' denotes the instruction tuning of DDT-LLaMA, while ``Tokenizer'' denotes the training of DDT.}
    \vspace{-1em}
    \begin{adjustbox}{width=0.5\textwidth}
    \begin{tabular}{lccc}
    \toprule
    \textbf{Hyper-parameters} & \textbf{MLLM-pt} & \textbf{MLLM-ft} & \textbf{Tokenizer}\\ \hline
    LLM init & LLama3-8B & MLLM-pt & - \\
    Optimizer & AdamW & AdamW & AdamW \\
    Optimizer param. & \multicolumn{2}{c}{$\beta_1=0.9,\beta_2=0.95,\epsilon=1\mathrm{e}{-6}$} & $\beta_1=0.9,\beta_2=0.99,\epsilon=1\mathrm{e}{-6}$ \\
    Peak LR & 1e-4 & 1e-5 & 1e-4 \\
    LR scheduler & Cosine & Cosine & Linear+Cosine \\
    Batch size & 1280 & 256/128 & 1024 \\
    Training Steps & 360K & 160K & 140K \\
    Warmup Steps & 5K & 2K & 5K \\
    Weight decay   & 0.05 & 0.05 & 0.0 \\
    Gradient clipping    & 1.0 & 1.0 & - \\
    Numerical precision    & bfloat16 & bfloat16 & bfloat16 \\
    Resource Usage    & 512 Ascend 910B & 256 Ascend 910B & 32 NVIDIA A800 \\
    Framework    & Megatron(TP=8) & Megatron(TP=8) & DDP \\
    \bottomrule
    \end{tabular}
    \end{adjustbox}
    \vspace{-1em}

\end{table}

\begin{figure*}[t]
\centering
% \vspace{-1em}
\includegraphics[width=1.0\textwidth]{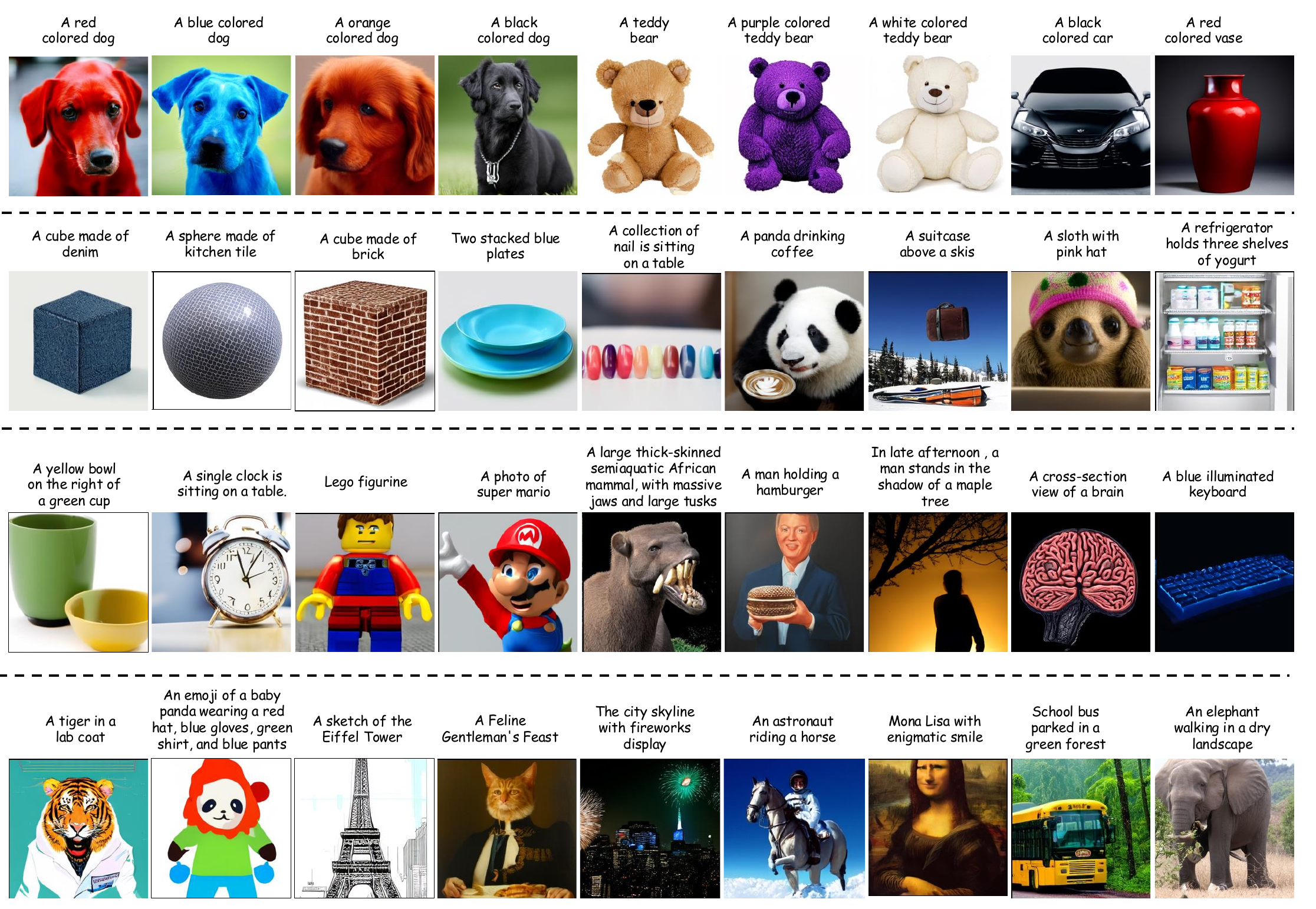}
\vspace{-2em}
\caption{More qualitative results of DDT-LLaMA text-to-image generation. (\textit{the supplement to Figure 3 in the main paper})}
\vspace{-1em}
\label{fig:t2i}
\end{figure*}

\begin{figure*}[t]
\centering
% \vspace{-1em}
\includegraphics[width=1.0\textwidth]{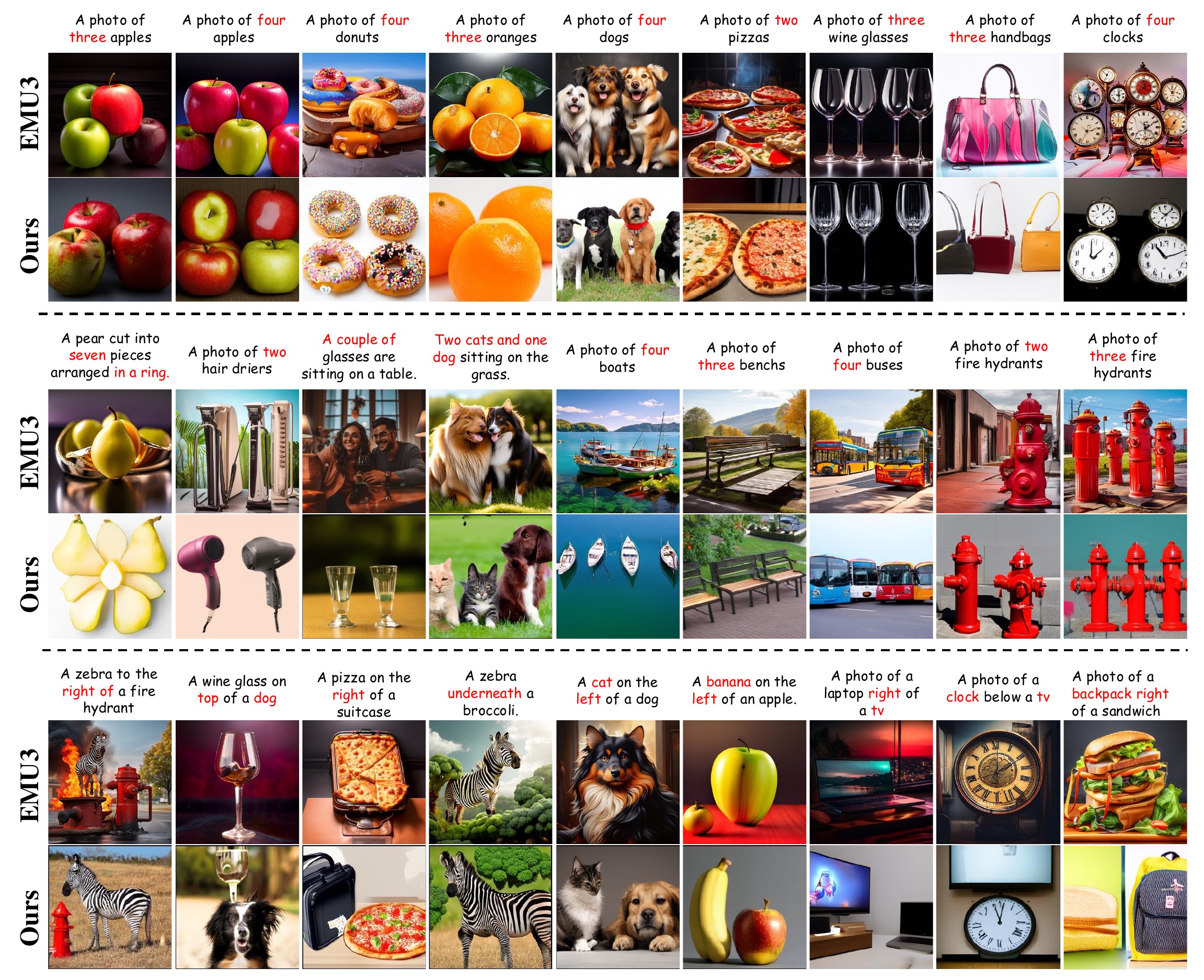}
\vspace{-2em}
\caption{ More qualitative comparison with EMU3 on T2I generation (PART-1). DDT-LLaMA can better respond to prompts related to counting, color, and position. (\textit{the supplement to Figure 4 in the main paper})}
\vspace{-1em}
\label{fig:emu3-1}
\end{figure*}

\begin{figure*}[t]
\centering
% \vspace{-1em}
\includegraphics[width=1.0\textwidth]{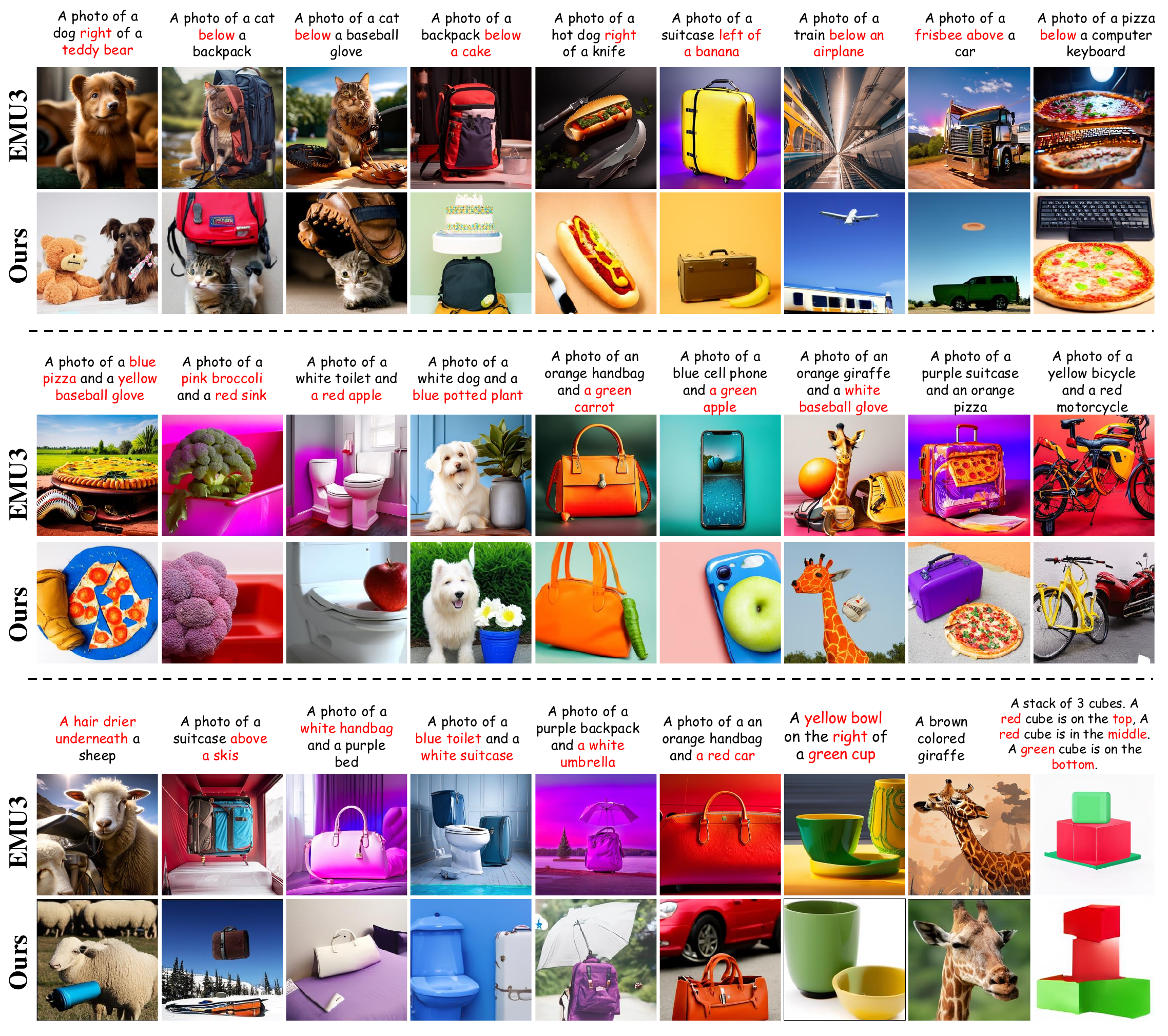}
\vspace{-2em}
\caption{ More qualitative comparison with EMU3 on T2I generation (PART-2). DDT-LLaMA can better respond to prompts related to counting, color, and position. (\textit{the supplement to Figure 4 in the main paper})}
\vspace{-1em}
\label{fig:emu3-2}
\end{figure*}

\begin{figure*}[t]
\centering
% \vspace{-1em}
\includegraphics[width=1.0\textwidth]{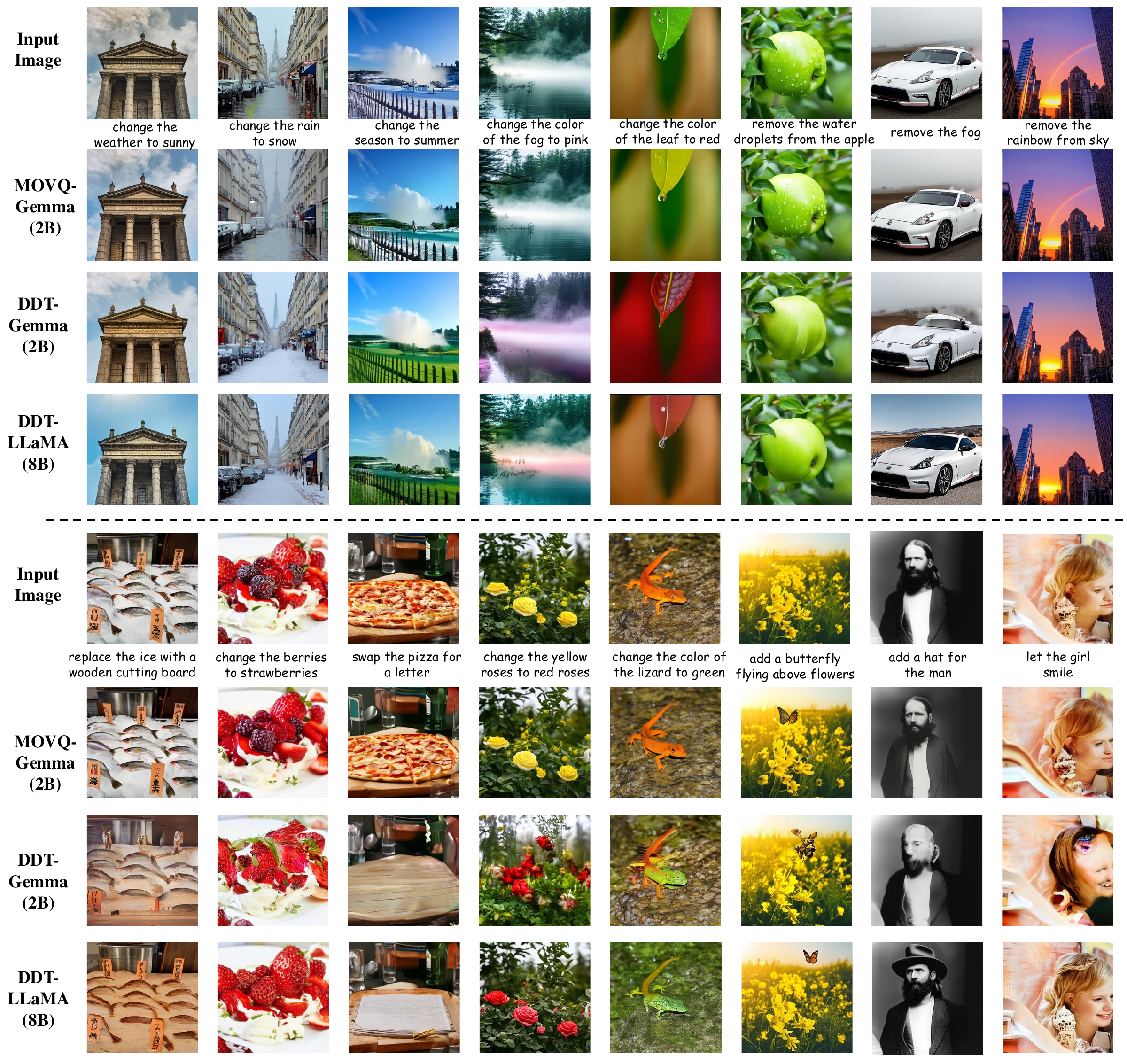}
\vspace{-2em}
\caption{Qualitative comparison of MOVQ-Gemma(2B), DDT-Gemma(2B), and DDT-LLaMA(8B) in the image editing task. In most instances, DDT-Gemma outperforms MOVQ-Gemma. Furthermore, DDT-LLaMA not only effectively comprehends and executes editing instructions accurately but also excels in preserving image fidelity. }
\vspace{-1em}
\label{fig:edit}
\end{figure*}

\begin{figure*}[t]
\centering
% \vspace{-1em}
\includegraphics[width=1.0\textwidth]{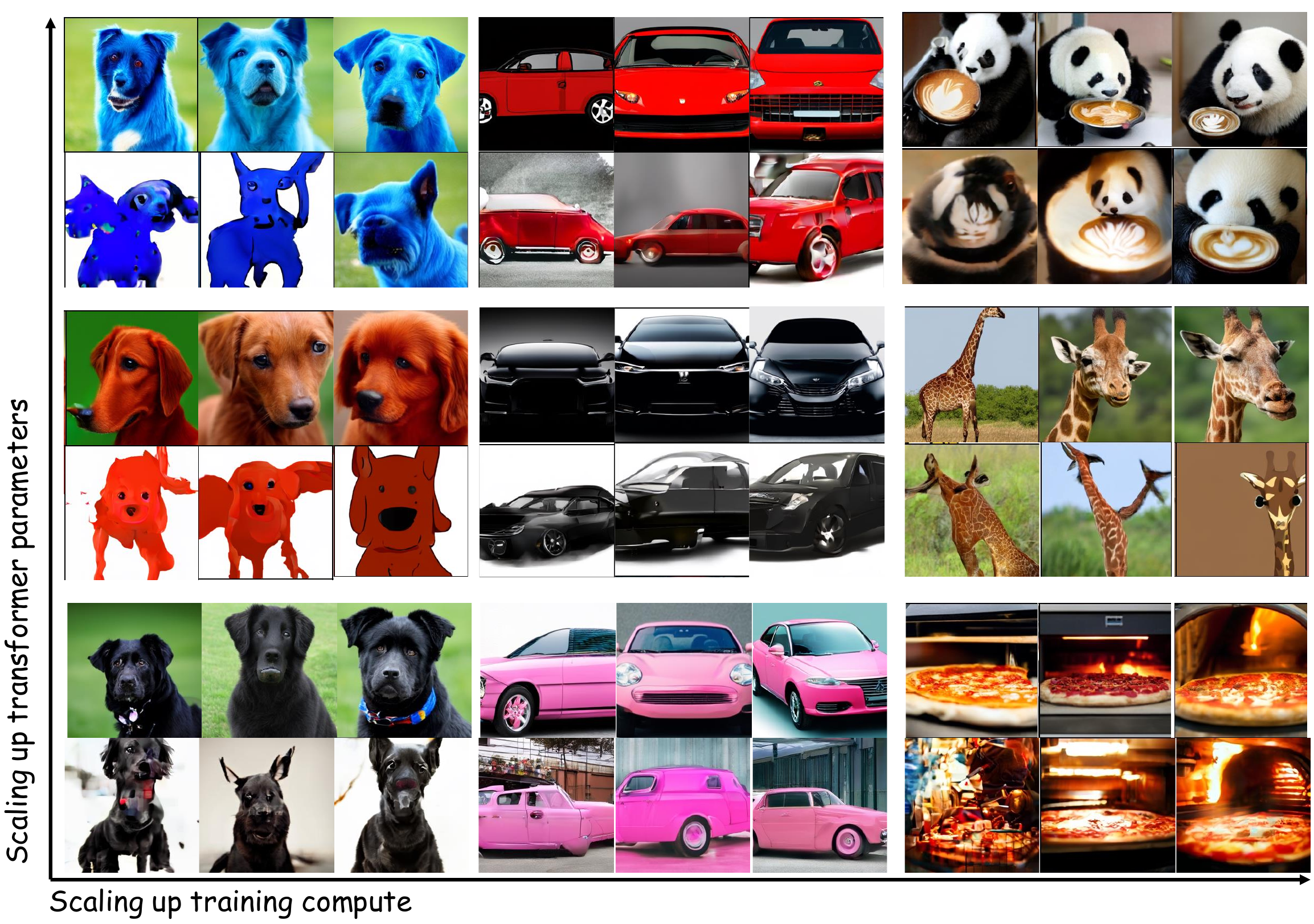}
\vspace{-2em}
\caption{ (More examples of text-to-image generation with different MLLM size (2B, 8B) and training compute (50\%, 75\%, 100\% of total tokens). \textit{the supplement to Figure 7 in Section 5.4.5 of the main paper})}
\vspace{-1em}
\label{fig:scaling}
\end{figure*}

\begin{figure}[t]
\centering
% \vspace{-1em}
\includegraphics[width=0.5\textwidth]{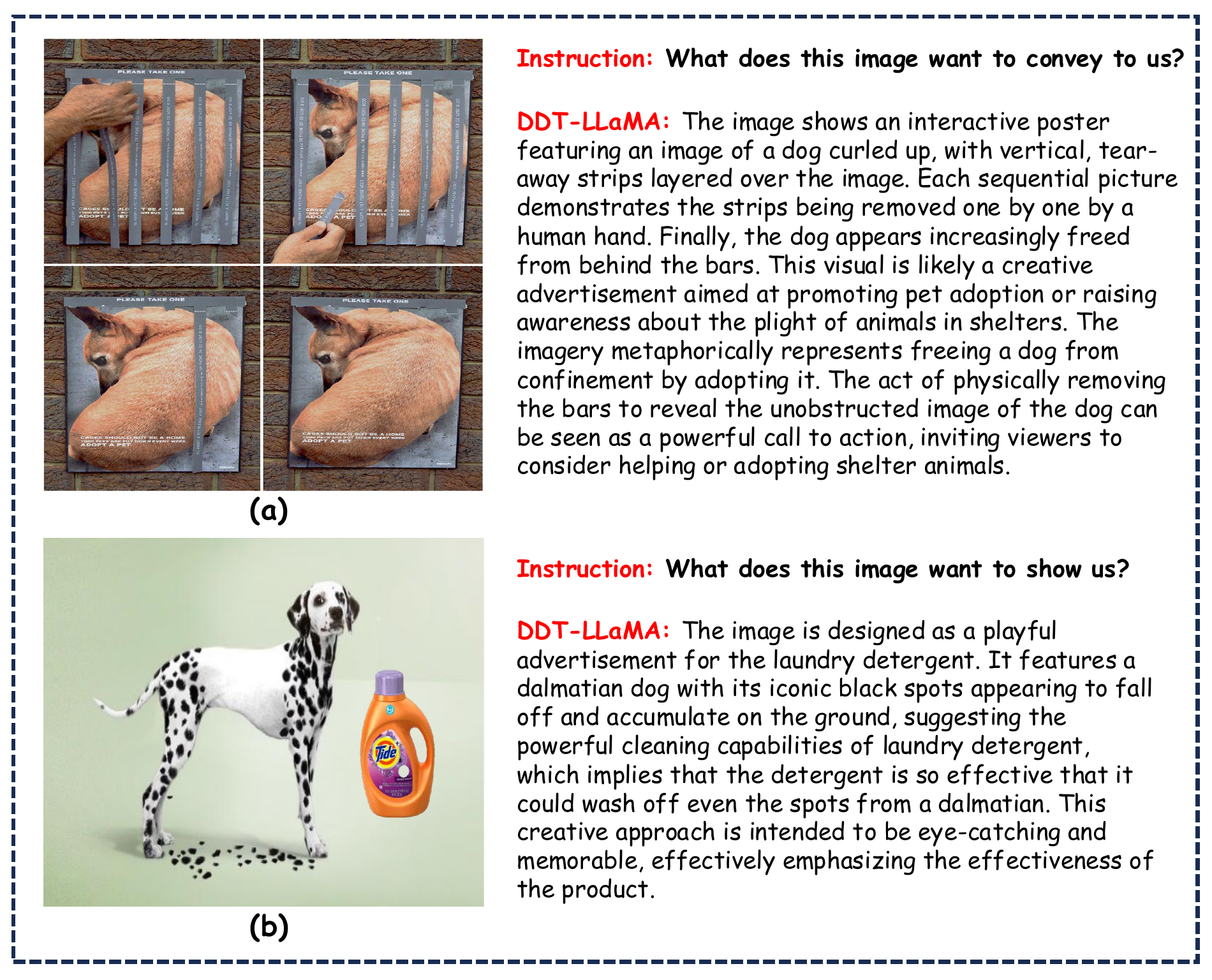}
\vspace{-2em}
\caption{Qualitative results of DDT-LLaMA visual comprehension. }
\vspace{-1em}
\label{fig:comp}
\end{figure}

\begin{figure}[t]
\centering
% \vspace{-1em}
\includegraphics[width=0.5\textwidth]{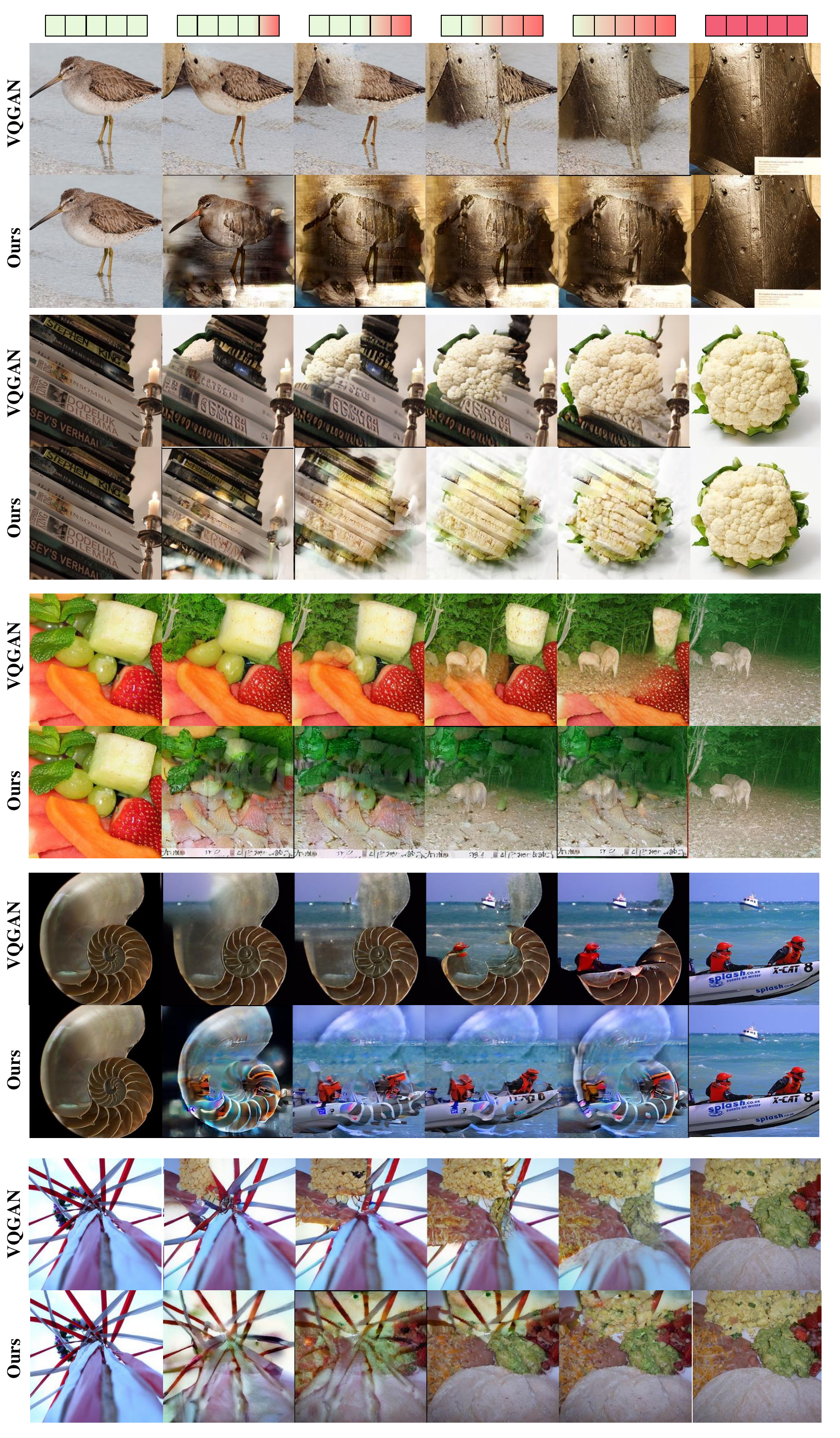}
\vspace{-2em}
\caption{More results of counterfactual interpolation with DDT tokens
and VQGAN tokens. (\textit{the supplement to Figure 6 in Section 5.4.2 of the main paper})}
\vspace{-1em}
\label{fig:interpolation}
\end{figure}

\begin{figure}[t]
\centering
% \vspace{-1em}
\includegraphics[width=0.5\textwidth]{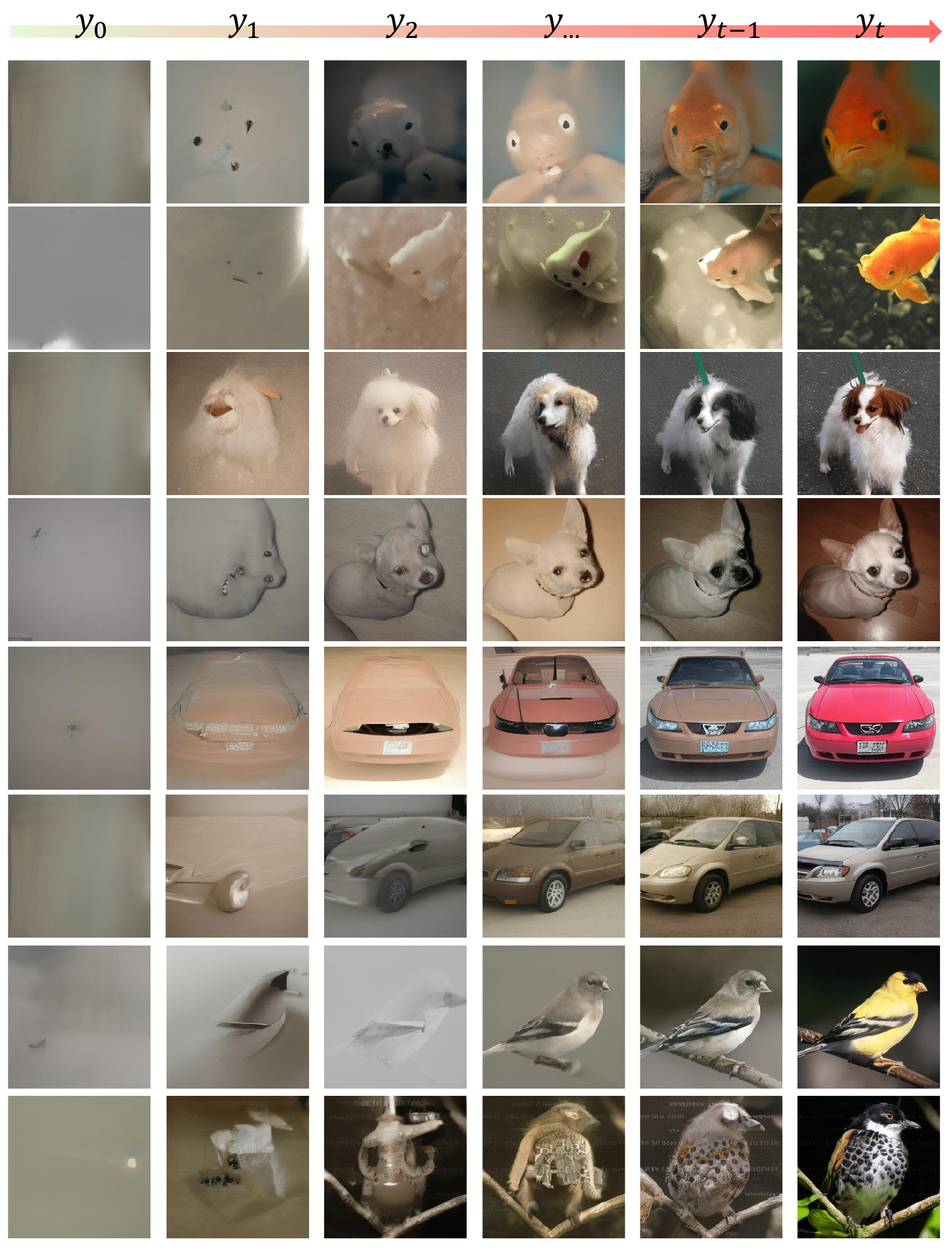}
\vspace{-2em}
\caption{More results of decoding images with an expanding subset of autoregressive-sampled DDT tokens (from 1 to $T=480$). $y_k$ denotes the number of sampled DDT tokens that are fed into the decoder for image generation, $1=y_0<y_1<y_2<...<y_{t-1}<y_t=480$. (\textit{the supplement to Figure 8 in Section 5.4.2 of the main paper})}
\vspace{-1em}
\label{fig:mask}
\end{figure}

\subsection{MLLM}
We initialize our MLLM from a pretrained LLM, specifically using the Llama3-8b model~\cite{dubey2024llama}, which has only undergone pretraining \textit{without instruction tuning}. 
Additionally, we expand its vocabulary by adding $|\mathcal{C}|=65,536$  visual codes and two extra special tokens (\texttt{[BOV]} and \texttt{[EOV]}).
Since both image and text are represented as discrete token IDs, we can use the cross-entropy to supervise the token prediction at each position for both modalities with a shared prediction head.
A shared head has proven more effective than using separate heads for each modality in training, enhancing the upper capabilities of the MLLM.
During inference, when generating content for a specific modality, tokens that do not fall in that modality's space should be masked. 
For example, when generating text, we mask the logits of the $65,536$ visual codes before sampling the text tokens; 
Similarly, when the \texttt{[BOV]} token indicates the beginning of image output, we mask the logits for $128,256$ text words before sampling the visual tokens. 
Moreover, we set $topk=50, topp=1.0$ for text token sampling and $topk=4096, topp=0.9$ for visual token sampling.
Besides, for text-to-image generation, during inference we use classifier-free guidance on the logits for autoregressive sampling in a manner similar to \cite{liu2024world, wang2024emu3}. We set the guidance scale to $8.0$.

During both pretraining and instruction-tuning, all parameters of the MLLM are \textbf{fully fine-tuned}. The hyper-parameters of both training stages for DDT-LLaMA are shown in \textcolor{red}{\textbf{Table~\ref{tab:hyper}}}.
The overall training is stable; we observed only one minor spike in the loss curve during pretraining. Following \cite{chowdhery2023palm}, we resume training from a checkpoint approximately 500 steps before the onset of the spike.

\section{Training Data \& Evaluation Details}

\subsection{Training Data}

\paragraph{Tokenizer Training.}
Our diffusion timestep tokenizer is trained on the training split of ImageNet~\cite{deng2009imagenet}, which comprises about 1.28 million images. Besides, each training image is center-cropped to a size of $256\times 256$. \textbf{The training was conducted on 32 NVIDIA A800 GPUs and lasted for nearly one weak.}

\paragraph{MLLM Pretraining.}
Our pretraining dataset, sourced from Laion~\cite{schuhmann2022laion}, consists of 200 million text-to-image pairs. 
Each pair includes an image accompanied by a brief, coarse-grained native caption and a detailed, fine-grained generated caption. 
For the generation of detailed captions, we employ ShareGPT-4V~\cite{chen2023sharegpt4v} to annotate over 400 million images sourced from Laion~\cite{schuhmann2022laion} and Coyo~\cite{coyo}.
Considering the propensity of ShareGPT-4V to generate captions that may not align with the images due to hallucinations, we utilize CLIP scores~\cite{radford2021learning} to filter all generated image-text pairs, retaining only those with the highest CLIP scores, totaling 200 million images.
For pretraining, we structure each pair in the format: ``\texttt{[BOS] <caption text> [BOV] <DDT tokens> [EOV] [EOS]}'' for pretraining, where \texttt{[BOS]} and \texttt{[EOS]} are the original special tokens from the text tokenizer,  \texttt{[BOV]} and \texttt{[EOV]} marking the start and the end of the vision input. 
Each image has a 60\% chance of being paired with the long caption and a 40\% probability of being paired with the short caption during training. 
Besides, each sample's caption has a 10\% probability of being dropped out.
Furthermore, to preserve the textual capabilities of MLLM, we supplement our dataset with purely textual data from Wikipedia and Pile~\cite{gao2020pile}, at a ratio of 10\%.
\textbf{The pre-training was conducted on 512 Ascend 910B NPUs and lasted for nearly two weaks.}

\paragraph{MLLM Instruction Tuning.}

During instruction tuning, we incorporate a variety of tasks, outlined as follows:
(1) Text-to-Image Generation: We employ datasets including ShareGPT4V caption~\cite{chen2023sharegpt4v},  ALLaVA~\cite{chen2024allava} and GRIT~\cite{chen2024allava}, utilizing a prompt template formatted as:``\texttt{[BOS] USER: <caption> Please Generate an image. ASSISTANT: [BOV] <DDT tokens> [EOV] [EOS]}''.

(2) Image editing:
We employ datasets such as InstructPix2Pix~\cite{brooks2023instructpix2pix} and Hive~\cite{zhang2024hive},  utilizing a prompt template formatted as:``\texttt{[BOS] USER: [BOV] <input DDT tokens> [EOV] <instruction> Please Generate an image. ASSISTANT: [BOV] <output DDT tokens> [EOV] [EOS]}''. Besides, considering the generally mediocre quality of existing image editing datasets, we also construct a batch of higher quality image editing data for integration into the training.

(3) Image caption \& VQA: We mainly leverage ShareGPT4V(-instruct)~\cite{chen2023sharegpt4v},  ALLaVA(-instruct)~\cite{chen2024allava} and select part of the held-in training datasets in InstructBlip~\cite{instructblip} for instruction-tuning. We follow the \cite{instructblip} to design the instruction templates.

\subsection{Evaluation Details}

\paragraph{Baseline Methods.}
For text-to-image generation tasks, we compare DDT-LLaMA with both diffusion-based T2I specialists and MLLM-based generalists. The diffusion-based T2I specialists include DALL-E 2~\cite{ramesh2022hierarchical}, SDv1.5~\cite{rombach2022high}, SDv2.1~\cite{rombach2022high}, SDXL~\cite{podell2023sdxl}, PixArt-alpha~\cite{chen2023pixartalphafasttrainingdiffusion}, DALL-E 3~\cite{betker2023improving}, and SD3~\cite{esser2024scaling}. For the MLLM-based generalists, we include comparisons with SEED-LLaMA~\cite{ge2023making}, LaVIT~\cite{jin2024unified}, Emu2-Gen~\cite{sun2024generative}, SEED-X~\cite{ge2024seed}, VILA-U~\cite{wu2024vila}, Lumina-mGPT~\cite{liu2024lumina}, and Emu3~\cite{wang2024emu3}. \textbf{\textit{(For EMU3, we report its results without prompt rewriting for fair comparison.}})

For image editing tasks, we compare DDT-LLaMA with both specialized image editing models and generalist MLLM-based models. The image editing specialists we evaluate include InsPix2Pix~\cite{brooks2023instructpix2pix}, MGIE~\cite{fu2023guiding}, and UltraEdit~\cite{zhao2024ultraedit}. Among the MLLM-based generalist models, we compare against GILL~\cite{koh2024generating}, Emu2-Gen~\cite{sun2024generative}, SEED-LLLaMA~\cite{ge2023making}, LaVIT~\cite{jin2024unified}, and SEED-X-Edit~\cite{ge2024seed}. \textbf{\textit{(We exclude MLLMs such as EMU3 because they lack image editing capabilities.}})

For visual comprehension and generation tasks, we compare DDT-LLaMA with specialized visual comprehension MLLMs, and MLLMs capable of both visual comprehension and generation. The specialized visual comprehension models include InstructBlip\cite{instructblip}, QWenVL-Chat~\cite{Qwen-VL}, LLaVA-1.5~\cite{liu2024improved}, mPLUG-Owl2~\cite{ye2024mplug}, ShareGPT4V~\cite{chen2023sharegpt4v}, LLaVA-1.6(HD)~\cite{liu2024llavanext}, and VILA~\cite{lin2023vila}. For models supporting both visual comprehension and generation, we compare DDT-LLaMA against Emu2-Chat~\cite{sun2024generative}, SEED-LLLaMA~\cite{ge2023making}, VILA-U~\cite{wu2024vila}, LaVIT~\cite{jin2024unified}, and Emu3~\cite{wang2024emu3}.

\paragraph{Evaluation Dataset.}
For text-to-image generation tasks, we conduct zero-shot evaluation on 3 benchmarks: GenEval~\cite{ghosh2024geneval}, T2I-CompBench~\cite{huang2023t2icompbench}, and DrawBench~\cite{saharia2022photorealistic}. GenEval contains 6 different subtasks of varying difficulty requiring various compositional skills, including \texttt{single object} (SingObj), \texttt{single object} (TwoObj), \texttt{counting}, \texttt{colors}, \texttt{position}, \texttt{color binding} (ColorAttri). 
And we adopt the metric proposed by ~\cite{ghosh2024geneval} for evaluation.
Each subtask is scored independently, and the overall score is calculated as the average of all six subtask scores. 
The T2I-CompBench suite encompasses six subtasks: \texttt{color}, \texttt{shape}, \texttt{texture}, \texttt{spatial}, \texttt{non-spatial}, and \texttt{complex} (complex compositions). Building on prior research, we employ the Blip-VQA score~\cite{li2022blipbootstrappinglanguageimagepretraining} to assess the color, shape, and texture subtasks. For spatial evaluation, we use the UniDet score~\cite{zhou2022simplemultidatasetdetection}; for non-spatial evaluations, the CLIP score~\cite{radford2021learning, hessel2022clipscorereferencefreeevaluationmetric}; and for complex compositions, the 3-in-1 Metric~\cite{huang2023t2icompbench}.
In terms of DrawBench, we leverage Clip text-visual feature similarity~\cite{radford2021learning} as the evaluation metric.

We also conduct zero-shot instruction-based image editing across three datasets: EVR~\cite{tan2019expressing}, MA5k~\cite{shi2021learning}, and MagicBrush~\cite{zhang2023magicbrush}. 
Following \cite{fu2023guiding}, for EVR and MagicBrush, we treat the standard pixel difference (L1) and visual feature similarity from the CLIP visual encoder (CVS) between generated images and ground-truth goals as the evaluation metrics. For MA5K, we utilize L1 and LPIPS~\cite{zhang2018unreasonable} as the evaluation metrics.

For visual comprehension tasks, we conduct zero-shot evaluation on a wide range of academic benchmarks, including image caption (NoCaps~\cite{agrawal2019nocaps}, Flickr30K~\cite{plummer2015flickr30k}), VQA (VQAv2~\cite{goyal2017making}, GQA~\cite{hudson2019gqa}, OKVQA~\cite{marino2019ok}, VizWiz~\cite{bigham2010vizwiz}), MLLM-oriented Comprehension Benchmarks (MME~\cite{fu2024mmecomprehensiveevaluationbenchmark}, SEEDBench~\cite{li2023seed}, POPE~\cite{li2023evaluatingobjecthallucinationlarge}).
(Note: In Table3 of the main paper, we use ``VQA'' to denote VQAv2). 
We employ CIDEr as the metric for image caption tasks, and VQA accuracy for VQA tasks.
% For MLLM-oriented Comprehension Benchmarks, we 严格follow对应数据集所提出的测试方法，and report perception score for MME, MCQ accuracy for SEEDBench, F1 score for POPE.
We employ the CIDEr metric to evaluate performance on image captioning tasks, while using VQA accuracy for the VQA datasets. Moreover, each MLLM-oriented benchmark is evaluated according to its specific prescribed methodologies, where we report the perception score for MME, MCQ accuracy for SEEDBench, and the F1 score for POPE.

\section{Additional Examples of T2I Generation}
In \textcolor{red}{\textbf{Figure~\ref{fig:t2i}}}, we present more qualitative examples of DDT-LLaMA on text-to-image generation tasks. 
DDT-LLaMA adeptly handles various types of instructions, including complex ones such as generating surreal images (\textit{e.g.}, ``A panda drinking coffee'', `` sloth with pink hat'')  and multi-condition combined prompts (\textit{e.g.}, ``An emoji of a baby panda wearing a red hat, blue gloves, green shirt, and blue pants''), to generate semantically-consistent images.

Furthermore, in \textcolor{red}{\textbf{Figure~\ref{fig:emu3-1}}} and \textcolor{red}{\textbf{Figure~\ref{fig:emu3-2}}}, we present a direct comparison between DDT-LLaMA and Emu3 across 54 prompts involving counting, color, and positioning. 
It is evident that Emu3 falls short in these areas:
(1) For counting-related prompts, EMU3 often generates images with \textit{an incorrect number of objects}.
(2) For prompts related to positioning, Emu3 frequently \textit{misplaces objects}, and sometimes \textit{only one of the objects} is generated.
(3) For color-related prompts, EMU3 often \textit{incorrectly assigns colors to the objects}, and it may also generate images where the arrangement or presence of objects is \textit{disordered} (``A photo of a purple suitcase and an orange pizza'' in Figure~\ref{fig:emu3-2}).
In contrast, DDT-LLaMA generates images that \textbf{more accurately reflect the desired object attributes (number and color) and adhere to the spatial specifications outlined in the prompts}.

\section{Additional Comparison on Image Editing}
As discussed in Section 5.4.4, we also employ Gemma2-2b~\cite{team2024gemma2} as the initial LLM and leverage both DDT tokens and MoVQ tokens for pretraining and instruction tuning, which we refer to as DDT-Gemma and MOVQ-Gemma, respectively.
In \textcolor{red}{\textbf{Figure~\ref{fig:edit}}}, we showcase a series of qualitative examples that compare MOVQ-Gemma (2B), DDT-Gemma (2B), and DDT-LLaMA (8B).  First, when comparing MOVQ-Gemma and DDT-Gemma, it is evident that in many editing cases,\textit{ MOVQ-Gemma often gives up editing}, typically returning the original image as the output. In contrast, DDT-Gemma exhibits a more robust comprehension of the editing instructions and delivers superior results in A/B tests, which indicates that \textbf{our recursive DDT tokens outperform spatial tokens in image editing tasks}.

Furthermore, DDT-Gemma sometimes faces problems with incomplete modifications or fails to maintain image fidelity in areas not targeted by the edits. For example, in the 7th case of Figure~\ref{fig:edit} (``remove the fog''), only the red exhaust behind the car is eliminated by DDT-Gemma. In the 12th case (``change the yellow roses to red roses''),  the shape of the roses is also inadvertently changed by DDT-Gemma. 
In contrast,\textbf{scaling up the backbone model from 2B to 8B significantly improves the editing performance}. We can see that DDT-LLaMA not only effectively comprehends the instructions to \textbf{accurately execute the editing},  but also excels at \textbf{preserving image fidelity}. This serves as evidence of the scaling-law properties of DDT tokens.

\section{Examples of Visual Comprehension}
In \textcolor{red}{\textbf{Figure~\ref{fig:comp}}}, we show some qualitative examples of visual comprehension. 
Although lack of pretrained encoders like CLIP~\cite{radford2021learning}, DDT-LLaMA can still effectively understand the visual semantics in the images and accurately infer the answer based on the given textual instruction.

\section{Additional Examples of In-depth Analysis}

\subsection{Counterfactual Interpolation}
\textcolor{red}{\textbf{Figure~\ref{fig:interpolation}}} shows more results of counterfactual interpolation of VQGAN tokens~\cite{esser2021taming} and DDT tokens. DDT tokens employ a disentangled representation to ensure that only the attributes represented by the substituted tokens vary in the generated counterfactuals, which \textbf{allows for a seamless semantic integration of the two images}.

\subsection{Decoding with a subset of DDT tokens}
In \textcolor{red}{\textbf{Figure~\ref{fig:mask}}}, we show more results demonstrating how autoregressive-sampled DDT tokens can be decoded into images in order. As the number of sampled tokens increases (from 1 to $T=480$), the image attributes are progressively reconstructed -- from fine details to the completion of coarse-grained contours and color information. This confirms that\textbf{ our DDT token sequence successfully decouples image attributes and possesses recursive properties}.

\subsection{Scaling Laws of DDT-based MLLM}
In \textcolor{red}{\textbf{Figure~\ref{fig:scaling}}}, we show more examples of text-to-image generation examples using two model sizes (Gemma 2B, LLama 8B~\cite{dubey2024llama}) at three different training stages (50\%, 75\%, and 100\% of total training tokens).  The enhancements in visual quality observed correspond with scaling laws, which indicate that \textbf{larger transformers trained on more comprehensive datasets tend to yield superior text-to-image performance}.

\section{Limitation and Future Work}

Our current tokenizer is trained solely on the ImageNet dataset~\cite{deng2009imagenet} at a resolution of 256x256 pixels, and it faces limitations in reconstructing open-domain images compared to baseline methods. For example, the EMU3 tokenizer MOVQ~\cite{zheng2022movq}, which is trained on a significantly larger dataset (from Laion and InternVID~\cite{wang2024internvidlargescalevideotextdataset}), achieves superior reconstruction performance than ours. As our 200M pretraining dataset filtered from an open-domain image dataset (\textit{i.e.}, Laion), the inadequate reconstructive capability of DDT restricts DDT-LLaMA's ability in text-to-image generation. This particularly impacts the aesthetic quality of the images generated by DDT-LLaMA, as illustrated in Figure~\ref{fig:t2i}, Figure~\ref{fig:emu3-1}, and Figure~\ref{fig:emu3-2}.

We are currently working on improving and scaling up the training of our DDT-tokenizer and the MLLM on a significantly larger dataset (about 500M images). 
\textit{\textbf{In the near future, we will release a more powerful version of DDT-LLaMA, along with a detailed technical report. Stay tuned!}}
Building on this foundation, we aim to further demonstrate that DDT-LLaMA is a significant approach for addressing visual-language tasks~\cite{Yu_2024_CVPR, Yu_2023_ICCV, yu2024anyedit, pan2024towards, li2023fine} and collaborative NLP tasks~\cite{hendrycks2021measuringmassivemultitasklanguage, zhu2022rosa, zhu2024graphclip, pan2024i3, pan2023self, wang2024mmluprorobustchallengingmultitask}. We also seek to extend the capabilities of DDT-LLaMA to support more vision-language tasks~\cite{lin2023tavt, lin2023exploring, lin2024non, linaction, huang2023vbenchcomprehensivebenchmarksuite, qiu2024step, li2023variational} such as video comprehension and video generation.

\end{document}